\documentclass{article}
\DeclareUnicodeCharacter{FF0C}{,}

\usepackage{arxiv}
\usepackage[square,numbers]{natbib}
\usepackage[utf8]{inputenc} % allow utf-8 input
\usepackage[T1]{fontenc}    % use 8-bit T1 fonts
\usepackage{hyperref}       % hyperlinks
\usepackage{url}            % simple URL typesetting
\usepackage{booktabs}       % professional-quality tables
\usepackage{amsfonts}       % blackboard math symbols
\usepackage{nicefrac}       % compact symbols for 1/2, etc.
\usepackage{microtype}      % microtypography
\usepackage{lipsum}		% Can be removed after putting your text content
\usepackage{graphicx}
\usepackage{doi}
\usepackage{verbatim}
\usepackage{mdframed}
\usepackage{fvextra}
\usepackage[english]{babel} 
\usepackage{csquotes}
\usepackage{tabularx}
\usepackage{pifont}
\usepackage{graphicx}
\usepackage{subcaption}
\usepackage{amsmath}
\usepackage{mdframed}
\usepackage{longtable}
\newcolumntype{P}[1]{>{\raggedright\arraybackslash}p{#1}}
\usepackage{comment}
\usepackage{csquotes}
\usepackage[table]{xcolor}
\usepackage{multirow}
\usepackage{array}
\usepackage{listings}
\usepackage{svg}
\usepackage{diagbox}
\usepackage[normalem]{ulem} % normalem prevents \emph from being underlined

\usepackage{fancyvrb}
\usepackage{listings}

\DefineVerbatimEnvironment{HighlightVerbatim}{Verbatim}
{fontsize=\fontsize{8pt}{10pt}\selectfont, commandchars=\\\{\},formatcom=\setlength{\parindent}{0pt}}

\usepackage{soul}
\usepackage{xcolor}

\usepackage{fancyvrb}

\title{What Level of Automation is "Good Enough"? A Benchmark of Large Language Models for Meta-Analysis Data Extraction}

\author{ 
	Lingbo Li \thanks{Corresponding author: L.Li5@massey.ac.nz} \\
	School of Mathematical and Computational Sciences\\
	Massey University\\
	Auckland, New Zealand \\
	\And
	\href{https://orcid.org/0000-0002-9124-2536}{\includegraphics[scale=0.06]{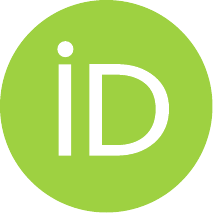}\hspace{1mm}Anuradha Mathrani} \\
	School of Mathematical and Computational Sciences\\
	Massey University\\
   Auckland, New Zealand\\
  	\And
\href{https://orcid.org/0000-0001-9416-1435}{\includegraphics[scale=0.06]{orcid.pdf}\hspace{1mm}Teo ~Susnjak} \\
	School of Mathematical and Computational Sciences\\
	Massey University\\
	Auckland, New Zealand \\
}

\renewcommand{\shorttitle}{...}

\begin{document}
\maketitle

\begin{abstract}
Automating data extraction from full-text randomised controlled trials (RCTs) for meta-analysis remains a significant challenge. This study evaluates the practical performance of three LLMs (Gemini-2.0-flash, Grok-3, GPT-4o-mini) across tasks involving statistical results, risk-of-bias assessments, and study-level characteristics in three medical domains: hypertension, diabetes, and orthopaedics. We tested four distinct prompting strategies (basic prompting, self-reflective prompting, model ensemble, and customised prompts) to determine how to improve extraction quality. All models demonstrate high precision but consistently suffer from poor recall by omitting key information. We found that customised prompts were the most effective, boosting recall by up to 15\%. Based on this analysis, we propose a three-tiered set of guidelines for using LLMs in data extraction, matching data types to appropriate levels of automation based on task complexity and risk. Our study offers practical advice for automating data extraction in real-world meta-analyses, balancing LLM efficiency with expert oversight through targeted, task-specific automation.
\end{abstract}

% keywords can be removed

\section{Introduction}
Meta-analysis is a gold standard in evidence-based medicine, combining quantitative findings from clinical trials to guide healthcare decisions \cite{cooper_research_2017, deeks2019meta}. Accurate data extraction is a foundational component in meta-analyses, as it directly affects the quality of findings and validity of the conclusions. Traditionally, this step has been highly manual and resource-intensive \cite{borah_analysis_2017}, requiring multiple reviewers to identify and double-check key details from each study \cite{higgins_cochrane_2019}. Errors at this stage are common and can compromise the accuracy of meta-analytic conclusions \cite{xu_validity_data_2022}. Furthermore, with the growing volume of the published research, the need for efficient, accurate and reliable automation methods capable of scalability, has become more critical \cite{marshall2019toward}.

Over the years, researchers have explored a range of automated data extraction approaches to alleviate the manual burden. Early efforts used rule-based systems with manually designed patterns to extract trial elements such as sample sizes and interventions \cite{kiritchenko2010exact, summerscales2011automatic}. While these showed that automated extraction was possible, they were constrained by the variability of reports, making them challenging to maintain or expand. To overcome such drawbacks, classical machine learning and specialised natural language processing techniques \cite{boyko_framework_2016, lorenz_automatic_2017, kiritchenko2010exact,michelson_automating_2014, lu_cheng_automated_2021} such as named entity recognition and supervised classification to extract PICO (Population, Intervention, Comparator, Outcome) elements are being employed. For example, BERT-based models, like those described by Mutinda et al. \cite{mutinda_automatic_2022}, offered better adaptability compared to rule-based systems. However, these earlier approaches needed annotated datasets \cite{kiritchenko2010exact, lu_cheng_automated_2021, boyko_framework_2016}, were often designed for narrow or specific objectives \cite{boyko_framework_2016, lorenz_automatic_2017}, and struggled with complex full-text documents \cite{kiritchenko2010exact, michelson_automating_2014, lu_cheng_automated_2021}. A recent literature review has noted that 84\% of extraction methods focused on abstracts, with only 25\% attempting to process full-text trial reports \cite{Schmidt2023dataextraction}. Furthermore, only a small proportion of projects produced widely accessible tools (approximately 8\% of methods had publicly available implementations) \cite{Schmidt2023dataextraction}, highlighting the difficulty of creating broadly applicable solutions. Despite these challenges, semi-automated tools like RobotReviewer \cite{marshall-etal-2017-automating} and AutoLit (https://about.nested-knowledge.com/) have emerged. RobotReviewer was among the first systems to address full-text data extraction and risk-of-bias assessment simultaneously. Though it aids reviewers by summarising participant details and methodological quality, it does not fully capture numerical outcomes essential for meta-analysis. AutoLit incorporates AI into the entire systematic review process, providing tools such as inclusion prediction and NLP-assisted data extraction. The platform’s value lies in its ability to streamline workflows, ensuring that extracted data flow directly into meta-analysis and visualisation components (the “Synthesis” module of the platform). Although AutoLit streamlines many aspects of the systematic review process, its outputs still require substantial human oversight \cite{nestedknowledge2023screening}. This is evident when extracting complex or numerical data from tables \cite{holub2021toward}.

More recently, large language models (LLMs) powered systems offer greater adaptability. Tools like MetaMate \cite{wang_metamate_2024} employ LLMs in a structured extraction pipeline with verification steps, achieving strong accuracy for participant and intervention data. In controlled settings, MetaMate attained F1 scores similar to those of human coders, including accurately parsing numerical expressions. However, it has not yet addressed full outcome extraction and has primarily been evaluated in the educational domain. The advent of general LLMs (such as ChatGPT \cite{openai_chatgpt}, Claude \cite{anthropic2023claude2}, and others) has introduced powerful new ways to automate data extraction. With LLMs' ability to understand context and perform tasks without extensive training from scientific texts, researchers are currently evaluating LLMs both as independent extractors and within human-in-the-loop workflows \cite{kartchner_zero-shot_2023, wang_metamate_2024, konet2024performance,yun_automatically_2024}. Studies show promising results: GPT-4 achieved 82\% accuracy extracting trial characteristics in biomedical RCTs \cite{schmidt2025exploringuselargelanguage}, and Claude successfully extracted binary outcome data with around 70–75\% accuracy \cite{yun_automatically_2024}. Konet et.al showed Claude 2 can achieve 96.2\% data elements correctness in 10 PDF articles \cite{konet2024performance}. However, their application to structured data extraction for meta-analysis has yet to be widely tested at the scale and complexity required for real-world use involving heterogeneous document types, especially when extracting detailed statistical information needed for meta-analyses. 

This study investigates the extent to which current LLMs can reliably extract the structured data from raw scientific papers required for automated meta-analysis (AMA), and further examines how extraction demands differ across statistical results, risk-of-bias assessments, and study-level categories of data, as well as how these data types impact extraction performance. We sought answers to these goals across three heterogeneous medical fields, namely, hypertension, diabetes, and orthopaedics, maximising the generalisability of our research. In our work, we evaluated the data extraction performance of three advanced models (Gemini-2.0-Flash, Grok-3, and GPT-4o-mini) against a human-annotated ground truth data, while exploring a variety of prompting and model-output aggregation strategies. Given the research momentum towards automating all parts of the meta-analysis process, our findings shed important light on the performance profile and the existing capabilities and limitations of current LLMs in real-world data extraction tasks for evidence synthesis purposes. Therefore, the contributions of this study are as follows: 
\begin{itemize}
    \item Comprehensive feasibility assessment: We provide the first in-depth benchmark of LLMs for full-text data extraction in meta-analysis across multiple clinical domains, information types, and task structures.
    \item Modular performance optimisation: We demonstrate that prompt specialisation and model output aggregation can yield distinct and complementary gains, establishing a modular strategy for improving extraction quality across heterogeneous AMA tasks.
    \item Methodologically robust evaluation protocol: We design a role-separated evaluation process that prevents LLMs from scoring their own outputs, using blinded, model-agnostic comparisons to ensure reliable and unbiased assessment of extraction quality.
    \item Task-specific automation guideline: Drawing on empirical performance patterns, we propose a three-tier classification of meta-analysis information types based on their suitability for automation and tolerance for error, offering practical guidelines on when to automate, when to review, and when human judgment remains essential.
\end{itemize}

\section{Methodology}

With LLMs at the centre of our methodology, our experimental design sought to answer the following questions: How reliable are generic LLM prompts at accurately and completely extracting relevant data from diverse fields covered by potential meta-analyses? If we follow up the original LLM prompt with a subsequent prompt, asking the LLM to revise and improve upon its initial data extraction, to what degree will the accuracy improve? Are we able to improve the overall accuracy of the data extraction if we combine the outputs of several different LLMs on the same task? And finally, if we move beyond generic prompts to more domain-specific and customised prompts for each targeted meta-analysis field, are we able to achieve greater accuracy?

\begin{figure}[h]
  \centering
  \includegraphics[width=\textwidth]{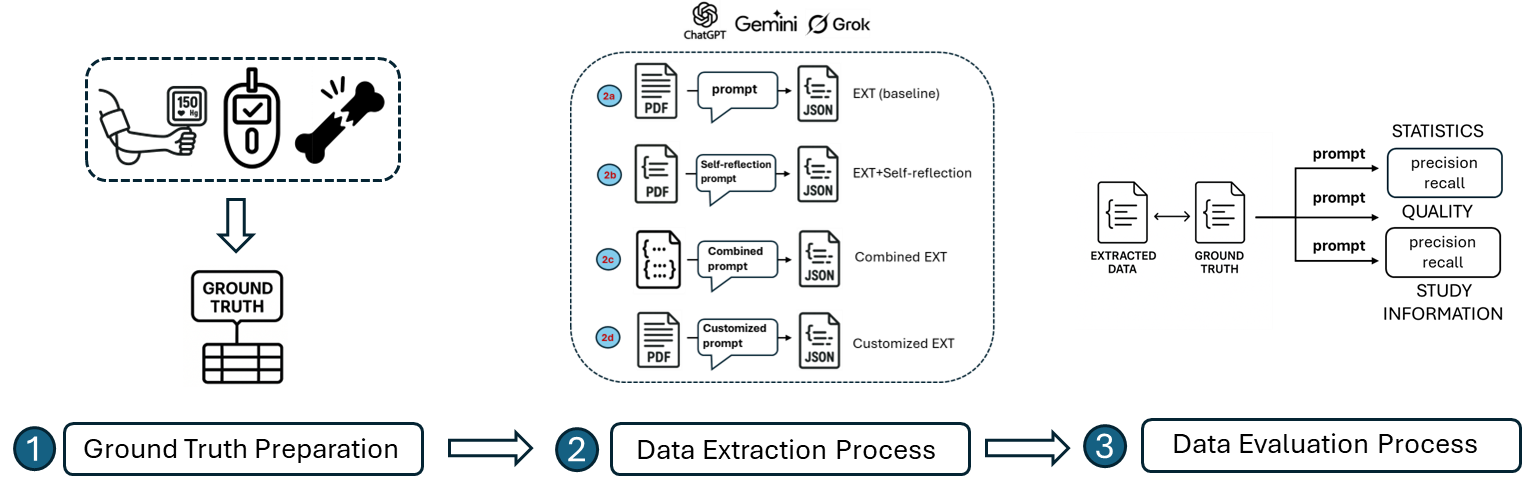}
  \caption{Overview of the whole workflow. Full-text RCTs were collected from published meta-analyses and annotated to construct a ground truth dataset, which served as the basis for evaluating and comparing multiple extraction methods}
  \label{fig:workflow}
\end{figure}

Figure \ref{fig:workflow} provides an overview of the empirical design, from ground truth construction to extraction and evaluation. The following subsections describe each stage of this process, beginning with the construction of the ground truth dataset for benchmarking.

\subsection{Study Selection and Ground Truth Preparation}
We sought to evaluate the accuracy and generalisability of our approach by replicating the datasets used by six diverse meta-analyses (MA1 - MA6) published between 2021-2025, whose underlying data we aimed to reproduce through our automated data extraction technique. We selected two studies from each of the three medical areas: hypertension (pharmacist-led care and diet interventions), diabetes (blood sugar control strategies and nutrition effects on insulin), and orthopaedics (bone fracture treatment and pharmacological impact on bone metabolism). These areas were chosen because they represent broad levels of complexity in how treatments and results are described in the papers. All were published in top-tier journals (Q1 or Q2). From each meta-analysis, we acquired and reviewed the full texts and identified the underlying RCTs studies, resulting in a total of 58 original manuscripts to serve as the testbed for automated data extraction. Table \ref{tab:meta_analyses} shows the characteristics of the six meta-analyses whose underlying data we attempted to reproduce via LLMs applied to the original 58 papers. 

To create the ground truth dataset, we manually extracted all the required data from the 58 RCTs and converted the dataset into a structured JSON format (Step 1 in Figure \ref{fig:workflow}). Importantly, we did not use data copied directly from the published meta-analyses, instead, we relied only on data directly extracted from the original RCT texts to ensure the ground truth was independently derived. We only extracted the specific measurements and outcomes that were actually used in the quantitative analyses of each meta-analysis, making sure our reference data matched what was actually synthesised. Two trained research staff independently reviewed and checked the JSON outputs against the analysis targets. We also drew on medical experts to validate our selection and interpretation of key variables. Any disagreements or unclear points were worked out through discussion. This expert-reviewed, agreed-upon dataset became our gold standard for testing how well the LLM-based extraction performed.

{
\begin{table}[h]
\centering
\caption{Characteristics of the Included Meta-Analyses}
\label{tab:meta_analyses}
%\small
\fontsize{8pt}{10pt}\selectfont
\begin{tabular}{p{0.5cm} p{1.5cm} p{1.5cm} p{1cm} p{5cm} p{1.5cm}}
\toprule
\textbf{No.} & \textbf{Author (Year)} & \textbf{Field} & \textbf{No. of RCTs} & \textbf{Primary Outcome(s)} & \textbf{Min/Max Tokens Used} \\
\midrule
MA1 & Matsumoto (2024) \cite{matsumoto_remote_2024} & Hypertension & 13 & Systolic Blood Pressure & 1033 / 3871\\
\addlinespace[0.5ex]
MA2 & Guo (2021) \cite{guo_effects_2021} & Hypertension & 10 & Systolic and Diastolic Blood Pressure \newline Anthropometric Measures (weight, BMI, waist circumference) \newline Metabolic Indicators (fasting glucose, total cholesterol, triglycerides, LDL-C, HDL-C) & 1549 / 3871\\
\addlinespace[0.5ex]
MA3 & Khalid (2023) \cite{khalid_effect_2023} & Diabetes mellitus & 10 & Lipid Profile Components (TC, TG, LDL, VLDL, HDL) & 1291 / 3097\\
\addlinespace[0.5ex]
MA4 & Yu (2025) \cite{yu_effect_2025} & Diabetes mellitus & 6 & HOMA-IR & 1549 / 2581\\
\addlinespace[0.5ex]
MA5 & Kim (2024) \cite{kim_effects_2024} & Orthopedic & 7 & BMD (femoral neck, total hip, lumbar spine) \newline BTM (CTX, P1NP, bone ALP, Osteocalcin) & 775 / 3613\\
\addlinespace[0.5ex]
MA6 & Oldrini (2022) \cite{oldrini_volar_2022} & Orthopedic & 12 & Functional Outcomes (DASH, PRWE, EQ-5D) at 3 and 12 months \newline  Range of Motion (flexion, extension, pronation, and supination) at 3, 6 and 12 months \newline Grip Strength (\% of the contralateral side) at 3, 6 and 12 months \newline Radiological Parameters (palar tilt, radial inclination, ulnar variance, step-off) in immediately post-surgery and at more than 3 months. & 1291 / 3355\\
\bottomrule
\end{tabular}
\end{table}
}

\subsection{LLM-based Extraction Process}
We evaluated the data extraction performance of three LLMs: \textbf{GPT-4o-mini} (OpenAI) \cite{openai2024gpt4omini}, \textbf{Gemini-2.0-flash} (Google DeepMind) \cite{google2024gemini2flash} and \textbf{Grok-3} (xAI) \cite{xai2024grok}, which we refer to henceforth simply as GPT, Gemini and Grok, respectively. We chose these models based on their reported \cite{windisch2024impact,schroeder2025llmloop,googlecloud2025geminidata,zhang2025grokqigong,microsoft2025grokazure} capabilities for various data extraction tasks, including information from medical literature, as well as evidence synthesis, together with practical considerations like cost and API availability. 

Each full-text RCT manuscript was processed directly by a chosen model as a PDF file without any preprocessing. Token analysis confirmed that all documents were well within each model’s maximum context window, ensuring the robustness of the experiments. As shown in Table \ref{tab:meta_analyses}, all PDF files were with the 4,000 token limit, enabling complete document input without truncation or splitting. Moreover, each model inference step was conducted with the temperature set to 0 to maximise deterministic outputs. The extraction followed a structured four-step prompting pipeline, and all models were required to return responses in a structured JSON format. All prompt templates used in this process are provided in Appendix. We designed four distinct data extraction strategies using all three LLMs depicted in Step 2 in Figure \ref{fig:workflow} follows:

\paragraph{Step 2a: Baseline Extraction (EXT)}
The initial experiment served as a baseline against which more refined approaches could be benchmarked. In this experiment, we designed a detailed but generic prompt which was medical-domain agnostic, with the instruction to extract the relevant study characteristics and outcome data. The objective was to extract relevant study characteristics and outcome data from each full-text RCT manuscript. For each extracted field, the model was also required to (i) assign a confidence level (high, medium, or low) and (ii) specify the source section within the full text from which the value was acquired. This was implemented as a safeguard against model hallucination and to promote traceability. The prompt clearly articulated the need for factual responses without invoking its internal inference or estimation. We refer to this baseline extraction output as \textbf{EXT}.

\paragraph{Step 2b: Self-Reflection and Revision (EXT+Self-reflection) }
In the second step, we instructed the model to review its own previous output through a process called \enquote{self-reflection} \cite{shinn2023selfreflection}. This process involves re-evaluating the initial extraction in order to identify and correct potential errors, as well as to revise confidence levels where appropriate. To enable this, the initial extraction results from  Step 2a were provided back to the same model, along with the original PDF document and a structured prompt requiring it to reflect on its first response. Self-reflection is a common approach for mitigating hallucination that considered effective in other tasks where models apply complex reasoning or multi-step thinking \cite{ji_etal_2023_towards,li_etal_2024_hindsight}. For meta-analysis, where important information is often scattered throughout a paper or buried in less obvious sections, the hypothesis was that this step may assist in locating omitted data points while correcting others. 
For each revision, the model was required to provide the original value, the new value, and a brief explanation of why it made the change. We refer to the outputs from this step \textbf{EXT+Self-reflection}. 

\paragraph{Step 2c: Combined Extraction (Combined EXT)}
Since all models have different capabilities and biases, our next hypothesis sought to determine if it is possible to improve the overall data extraction accuracy by combining the baseline responses of all three LLMs. The approach is based on ensemble learning theory, which states that combining outputs from different models, especially when they have complementary strengths, can enhance overall performance \cite{Dietterich2000ensemble, dong2020survey}. Recent studies have confirmed that ensemble methods combining outputs from multiple LLMs can better extract biomedical concepts from clinical text and improve tasks like identifying medical terms \cite{li2024ensemble, zhang2022biomedical}. 
We forwarded the baseline EXT outputs from GPT, Gemini, and Grok to a merging process. The merging process was performed by another LLM (Gemini), which was guided by a structured prompt containing rule-based instructions. The model was instructed to strictly follow the rules instead of creating new data points or making inferences. The merging rules followed a hierarchical logic: (1) If two models agreed on a value and the third differed, the majority value was retained; (2) If all three values were different, the value associated with the highest confidence score was selected (when available); (3) In the absence of confidence scores, the most complete and internally consistent value was chosen, based on predefined criteria such as field completeness and conformity to expected data types; (4) For nested fields (e.g., outcome measures or participant details), the same hierarchical approach was applied recursively. The prompt strictly prohibited any rewording, subjective commentary, or fabrication of data. Importantly, the merging LLM operated solely on the EXT outputs without access to the model identities or the original document content. This ensured that the merging process remained decoupled from the original extraction logic and unbiased by the specific characteristics of any single model. We call the final merged outputs \textbf{Combined EXT}.

\paragraph{Step 2d: Customised Extraction (customised EXT)}
Given the proclivity of LLMs to display responses in highly variable levels of accuracy and quality depending on the detail and structure of the prompts, we hypothesised that accuracy gains might be achievable by tailoring the prompt to the specific medical domain relevant to the study. Therefore, we created a set of domain-specific prompts, customised to match the topic of each meta-analysis (e.g., hypertension, diabetes, orthopaedics). These prompts were customised to reflect the specific outcomes and variables prioritised by the authors of each meta-analysis. For example, when the prompt primed the LLM to assume expertise in a field like \enquote{orthopedic and metabolic bone disease}, the LLM was explicitly instructed to attend to the domain-relevant data extraction with the guidance to: \enquote{\textit{Focus on these outcomes: Bone Mineral Density (femoral neck, total hip, lumbar spine) and Bone Turnover Markers (CTX, P1NP, BONE ALP, Osteocalcin)}}.
The customised prompts were applied across the same three LLMs (GPT, Gemini and Grok), and the outputs we call as \textbf{Customised EXT}. 

\subsection{Evaluation}
All extraction outputs (EXT, EXT+Self-reflection, Combined EXT, and customised EXT) were evaluated against the ground truth at the field-level. To reflect how extracted data is used in real-world meta-analyses, we organised the evaluation into three functional categories: 
(1) \textbf{statistical results}, which included sample sizes, interventions or comparison details, different outcomes including mean, standard deviation, effect sizes, confidence intervals, p-values, adverse events and dropouts; 
(2) \textbf{quality assessment}, covering risk of bias domains such as randomisation, allocation concealment, and blinding; and 
(3) \textbf{basic study information}, including basic study information such as title, author, years, population characteristics and eligibility criteria. These categories were derived from the structure of the model-generated JSON outputs. The ground truth dataset was organised into the same three categories, allowing direct comparison between model outputs.

\subsubsection{LLM-based Evaluation and Human Validation}
We employed the LLM (Gemini) to performed file-level evaluation to efficiently assess extraction quality. The evaluation was designed to be unbiased and reliable by structuring it as a series of simple comparisons between extracted outputs and their corresponding ground truth values, formatted as JSON key–value pairs. To ensure objectivity, Gemini received only the extracted value and the matching ground truth for each field. It did not have access to the original prompt, full document, or information about which model had produced the extraction. This setup was specifically intended to prevent evaluation bias, such as a model validating its own output, and to ensure that the assessment was based solely on semantic equivalence between extracted and reference values. For each field, the model was instructed to assign one of three evaluation labels: (1) \textbf{Correct}: The extracted value matched the ground truth in meaning. (2) \textbf{Missing}: The ground truth included a value, but the model failed to extract it. (3) \textbf{Hallucinated}: The model extracted a value that was not present or justifiable in the original text. For hallucinated fields, we implemented a secondary classification to better understand the nature of the error. These were broken down into: (1) \textit{Incorrect value}, where the extracted field was factually wrong; (2) \textit{Incorrect unit}, where the number was right but the unit was wrong; and (3) \textit{Overgeneralized}, where the extracted information information was too vague, non-specific, or lacked necessary detail/context. All evaluations used a temperature of 0.0 to ensure guarantee deterministic outputs. This approach enabled scalable, reproducible, and fine-grained assessment of extraction performance across thousands of data fields.

To check whether this LLM-based evaluation was reliable, we conducted a blinded manual review of 900 sample fields. Using stratified random sampling, we picked up to 30 fields from each combination of model (GPT, Gemini, Grok), extraction method (EXT, EXT+Self-reflection, Combined EXT, customised EXT), and data category (statistics, quality, and information). Two independent reviewers, who did not know the LLM-assigned labels or each other's responses, labelled each field as Correct, Hallucinated, or Missing by comparing it to the ground truth. Agreement between human reviewers and LLM-assigned labels was 96.09\%, and agreement between the two human reviewers was very high (Cohen's $\kappa = 0.987$), showing that the automated evaluation approach offers a reliable and efficient alternative to manual labelling for large-scale evidence extraction tasks.

\subsubsection{Performance Assessment}
We calculated \textbf{precision} and \textbf{recall} at the element level to measure how well the data extraction worked. Precision shows what proportion of extracted elements were correct, while recall shows what proportion of relevant elements from the ground truth were successfully found. We calculated these measures separately for each category and extraction method, giving us complementary views of how accurate and complete the extraction was. Then, we compared both the effectiveness of different extraction strategies and how well different models performed. A non-parametric Friedman test was conducted across task categories to assess whether the observed differences in performance were statistically significant, with a \textit{post-hoc} Nemenyi test used to identify differences. Based on this comparison, we identified the method with the best overall performance and then used it as a foundation for comparing models. Within this method, we ranked models by their recall scores, and calculated their \textbf{mean rank} to provide a summary measure of extraction quality.

\section{Results}
We present the evaluation results from three different angles. First, we look at overall performance, comparing different methods and models and examining how they interact with each other. Next, we evaluate performance differences across individual meta-analysis datasets to identify any trends specific to particular medical areas. Finally, we examine the distribution of different types of data extraction errors, both overall and in relation to specific fields and models. 

\subsection{Overall Performance}
Table \ref{tab:total_results} shows precision and recall for each method-model combination, where data extraction was evaluated and averaged across all 58 papers and broken down by different data types (i.e. statistics results, quality assessment and basic study information). Overall, extraction performance varied substantially by field type, especially in recall, while being relatively high in precision. Study information fields were generally extracted most completely (recall ranging from 0.52 to 0.84), followed by quality assessments (recall ranging from 0.7 to 0.78), while statistical results proved to be the most challenging (recall ranging from 0.21 to 0.76). The table also shows that the customised EXT approach outperformed others across all three categories of data type extractions by a clear margin.

{
\begin{table}[htbp]
\centering
\caption{Overall Performance for different extraction approaches across models}
\label{tab:total_results}
%\small
\fontsize{7pt}{9pt}\selectfont
\setlength{\tabcolsep}{5pt} 
\begin{tabular}{ccccc}
\toprule
\textbf{Approach} & \textbf{Model} & \textbf{Precision} & \textbf{Recall} & \textbf{Recall Rank} \\
\hline\hline
\multicolumn{5}{l}{\textit{\textbf{Statistics Results}}} \\ 
\hline\hline
\multirow{3}{*}{\centering EXT (Baseline)}    
& GPT  & 0.856 & 0.214 & \diagbox{}{} \\
& Gemini  & 0.911 & \textbf{0.445} & \diagbox{}{} \\
& Grok  & \textbf{0.939} & 0.389 & \diagbox{}{} \\
\midrule
\multirow{3}{*}{\centering EXT+Self-reflection}  
& GPT   & 0.854 & 0.231 & \diagbox{}{} \\
& Gemini   & 0.905 & 0.441 & \diagbox{}{} \\
& Grok  & \textbf{0.928} & \textbf{0.464} & \diagbox{}{} \\
\midrule
{\centering Combined EXT}          
& \diagbox{}{}  & 0.906 & 0.430 &  \diagbox{}{} \\
\midrule
\multirow{3}{*}{\centering Customised EXT}       
& GPT  & 0.814 & 0.381 &  3.0\\
& Gemini  & \textbf{0.952} & \textbf{0.760} & 1.0  \\
& Grok   & 0.921 & 0.697 &  2.0 \\
\hline\hline
\multicolumn{5}{l}{\textit{\textbf{Quality Assessment Results}}} \\ 
\hline\hline
\multirow{3}{*}{\centering EXT (Baseline)}    
& GPT  & 0.866 & 0.730 &  \diagbox{}{}\\
& Gemini & 0.914 & 0.702 & \diagbox{}{} \\
& Grok  & \textbf{0.936} & \textbf{0.760} & \diagbox{}{} \\
\midrule
\multirow{3}{*}{\centering EXT+Self-reflection}  
& GPT  & 0.878 & 0.744 & \diagbox{}{} \\
& Gemini  & 0.928 & 0.739 & \diagbox{}{} \\
& Grok  & \textbf{0.933} & \textbf{0.763} & \diagbox{}{} \\
\midrule
{\centering Combined EXT}          
& \diagbox{}{}  & 0.927 & 0.759 & \diagbox{}{} \\
\midrule
\multirow{3}{*}{\centering Customised EXT}       
& GPT  & 0.863 & 0.730 &  2.0 \\
& Gemini  & 0.920 & 0.727 & 3.0 \\
& Grok  & \textbf{0.953} & \textbf{0.782} &  1.0 \\
\hline\hline
\multicolumn{5}{l}{\textit{\textbf{Study Information Results}}} \\
\hline\hline
\multirow{3}{*}{\centering EXT (Baseline)}    
& GPT  & 0.770 & 0.521 & \diagbox{}{} \\
& Gemini  & 0.870 & 0.631 &  \diagbox{}{}\\
& Grok  & \textbf{0.913} & \textbf{0.754} &  \diagbox{}{}\\
\midrule
\multirow{3}{*}{\centering EXT+Self-reflection}  
& GPT  & 0.764 & 0.523 & \diagbox{}{} \\
& Gemini  & 0.864 & 0.646 & \diagbox{}{} \\
& Grok  & \textbf{0.931} & \textbf{0.756} &  \diagbox{}{}\\
\midrule
{\centering Combined EXT}          
& \diagbox{}{}  & 0.887 & 0.705 &  \diagbox{}{} \\
\midrule
\multirow{3}{*}{\centering Customised EXT}       
& GPT  & 0.745 & 0.760 & 3.0 \\
& Gemini  & 0.845 & 0.808 &  2.0\\
& Grok  & \textbf{0.887} & \textbf{0.841} & 1.0 \\
\bottomrule
\end{tabular}
\end{table}
}

\subsubsection{Extraction Method Comparison} \label{sec:methods comparison}
\begin{figure}[htbp]
    \centering
    \includegraphics[width=0.75\textwidth]{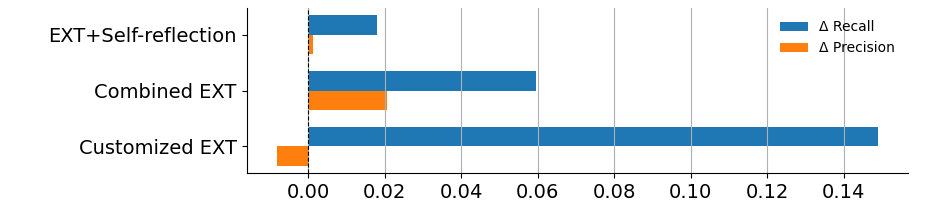}
    \caption{Average performance change ($\Delta$ precision and $\Delta$ recall) of three extraction strategies relative to the EXT baseline}
    \label{fig:methods}
\end{figure}

\begin{figure}[htbp]
    \centering
    \includegraphics[width=0.75\textwidth]{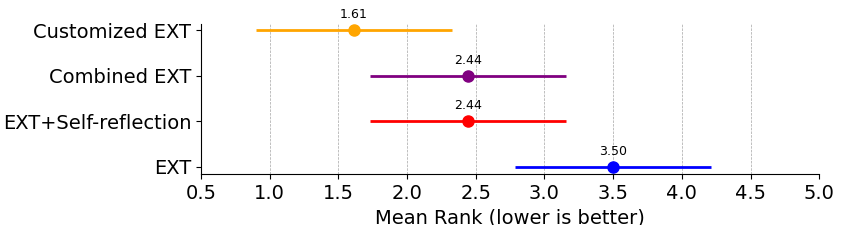}
    \caption{Friedman–Nemenyi critical difference (CD)  graph based on mean rank in recall. If the two horizontal line segments in the figure do not overlap, it signifies a significant performance difference between the two methods}
    \label{fig:fridedman}
\end{figure}

Here we quantify the degree to which each of the extraction method variants improved over the baseline EXT. Figure \ref{fig:methods} summarises the average change over the EXT in recall and precision for each method. Customised EXT delivered the biggest recall increase, with an average gain of 14.8\% across data categories and models. This came with a tiny precision drop (-0.8\%), showing a controlled trade-off between finding more relevant content and slight over-extraction. Combined EXT offered the most balanced results, achieving a solid recall gain (5.9\%) plus a small precision increase (2.0\%). This suggests that combining multiple models can improve both completeness and accuracy in a stable way.
Self-reflection EXT showed only minor improvements in both recall (1.8\%) and precision (0.1\%), suggesting that while instructing models to review and revise their responses may lead to more reliable outputs, it does not significantly aid with retrieving additional relevant information. A Friedman test confirmed that the recall differences between methods were statistically significant ($\chi^2$(3) = 9.81, \textit{p} = 0.0203), showing that an extraction strategy does have an effect on retrieval performance. 
The follow-up Nemenyi test found that EXT was significantly outperformed by customised EXT, though the difference between EXT and Combined EXT was not statistically significant (Figure \ref{fig:fridedman}). Combined EXT and EXT+Self-reflection showed similar overall recall performance, with matching mean ranks (2.44), suggesting their extraction effectiveness is nearly identical. 
Mean ranks were used to summarise how each extraction method performed across all model-task combinations. For each combination, methods were ranked based on their recall scores, with higher recall receiving better (i.e. lower) ranks. These ranks were then averaged to produce a mean rank for each method.

\subsubsection{LLM Model Performance Comparison}
To compare the capabilities of the LLM models for the data extraction task fairly without confounding factors from different extraction methods, we used customised EXT (which performed best overall) as the shared extraction baseline. For each of the three task categories, we ranked the models by recall, which are reported in Table \ref{tab:total_results}. Then, we further averaged each model's recall ranks across the three task categories to give mean ranks. The resulting mean ranks (although not shown in Table \ref{tab:total_results}) revealed that Grok achieved the best overall performance (1.3), followed by Gemini (2.0) and GPT (2.7). 

For the extraction of the statistical results category data, Gemini performed best (average rank 1.0), showing its strength with numerical data. Grok came second (2.0), while GPT lagged behind (3.0), suggesting smaller models struggle more with numerical content. For quality assessment, Grok ranked highest, GPT came second, Gemini came third. Quality assessment fields often use less structured, more subjective language that sometimes spans multiple sentences, making them harder to extract reliably. While all models showed some variability here, Grok indicated stronger capabilities at handling loosely structured content. For study information data extraction (e.g. structured metadata like titles, authors, years, population details, eligibility criteria), Grok again performed best (1.0), followed by Gemini (2.0) and GPT (3.0). This data category was evidently less challenging for all models, presumably due to this type of data tending to be represented using predictable wording and formatting. Grok depicted the best performance by significant margins,  perhaps suggesting that, with it being the newest model, there are increased data capture capabilities in the more advanced models on these types of tasks. Across all data type categories, Grok displayed the most reliable data extraction performance, Gemini excelled with complex or variable fields, while GPT lagged repeatedly behind the others. This confirms that model differences meaningfully affect extraction quality, especially when content varies in structure and clarity.

\subsubsection{Comparison of Method–Model Combinations}
All three models improved their recall when using the Customised EXT method (Table \ref{tab:model-performance}). Gemini realised the biggest boost (17.2\%), followed by Grok (13.9\%) and GPT (13.5\%). In contrast, EXT+Self-reflection produced only small recall changes, with modest increases for Grok (2.7\%), Gemini (1.6\%), and GPT (1.1\%). Combined EXT showed strong improvements for GPT (14.3\%) and a modest gain for Gemini (3.8\%), but slightly decreased Grok's performance (-0.3\%). Precision changes were generally small without clear patterns, though GPT did exhibit a notable precision increase (7.6\%) with the Combined EXT method. These results indicate that generating more tailored and customised data extraction prompts can effectively improve recall, while strategies that leverage ensemble-based solutions for data extraction can provide balanced improvements in both recall and precision for certain models like GPT.

{
\begin{table}[htbp]
    \centering
    \caption{Comparison of Model Performance Across Different Methods}
    \label{tab:model-performance}
    \fontsize{8pt}{10pt}\selectfont
    \begin{tabular}{lccc}
        \toprule
        \textbf{Model} & \textbf{Method} & \textbf{$\Delta$Recall} & \textbf{$\Delta$Precision} \\
        \midrule
        \multirow{3}{*}{GPT} 
        & EXT+Self-reflection & 0.011 & 0.001 \\
        & Combined EXT & \textbf{0.143} & \textbf{0.076} \\
        & Customised EXT & 0.135 & -0.023 \\
        \midrule
        \multirow{3}{*}{Gemini} 
        & EXT+Self-reflection & 0.016 & 0.001 \\
        & Combined EXT & 0.038 & \textbf{0.008} \\
        & Customised EXT & \textbf{0.172} & 0.007 \\
        \midrule
        \multirow{3}{*}{Grok} 
        & EXT+Self-reflection & 0.027 & \textbf{0.001} \\
        & Combined EXT & -0.003 & -0.023 \\
        & Customised EXT & \textbf{0.139} & -0.009 \\
        \bottomrule
    \end{tabular}
\end{table}
}

\subsection{Performance Across Meta-analyses}
Previous sections examined data extraction performance at the level of overall averages by data types, here we provide a deeper and more granular analysis, examining how different extraction methods performed across the six meta-analyses RCTs' datasets (MA1 -- MA6) as well as data types, in order to surface the existence of performance variabilities. Each meta-analysis RCTs' dataset varied in structure and field density, offering a useful perspective on the capability of each method. 

\subsubsection{Data Extraction Method Comparison}
\begin{figure}[htbp]
    \centering
    \includegraphics[width=\textwidth]{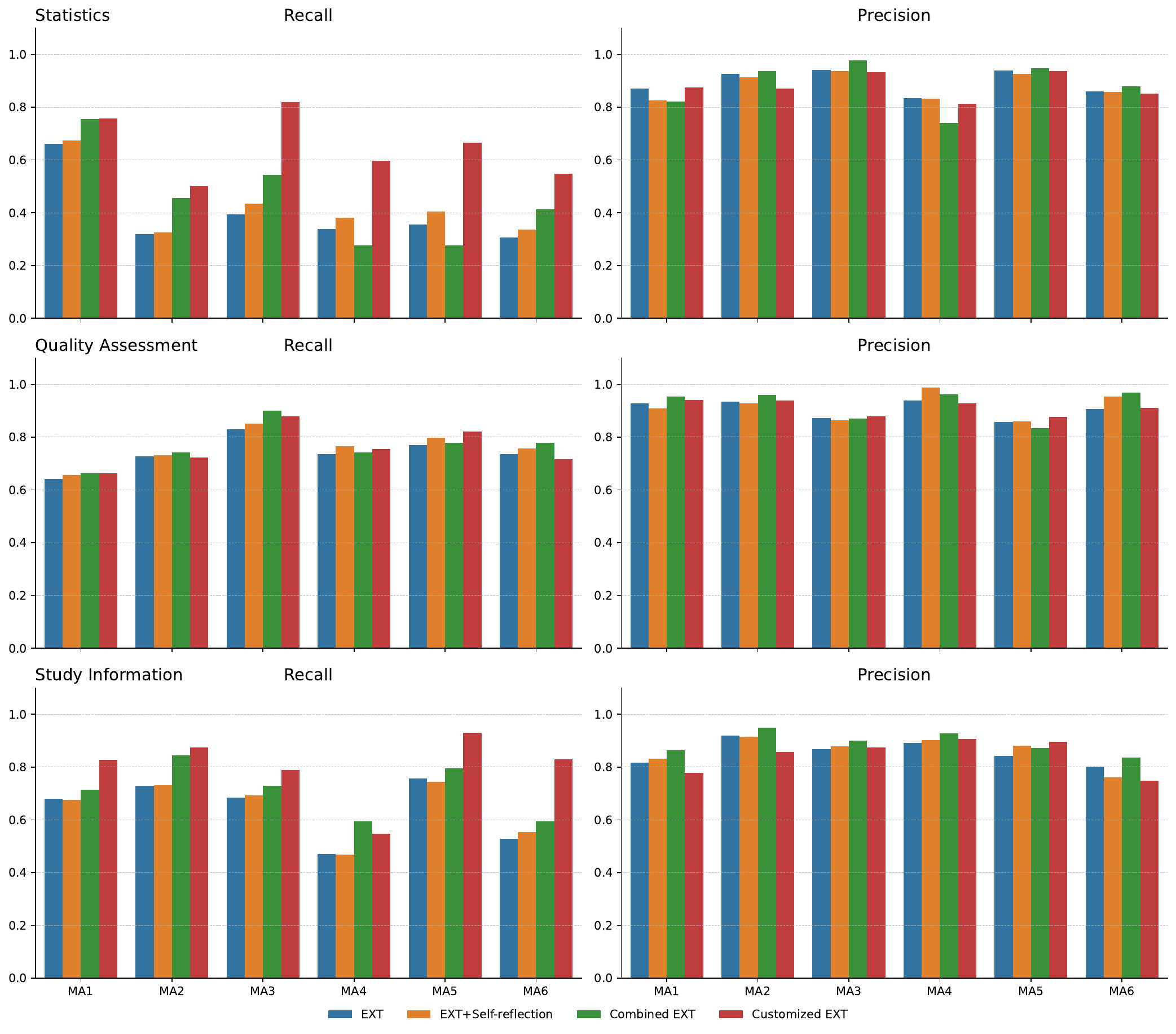}
    \caption{Methods Comparison Across Meta-analyses}
    \label{fig:cate_per}
\end{figure}

As shown in Figure \ref{fig:cate_per}, the performance of the different extraction methods varied significantly across different MA datasets. For statistical data extraction, methods struggled with MA datasets like MA3, MA4, MA5, and MA6, which contained high levels of specialised medical terms (e.g. lipid profiles and bone turnover markers) and complex structures such as multi-level groupings. In these challenging cases, Customised EXT achieved recall gains over 20\% compared to baseline methods, likely because it was already primed in the prompt for the specialised context and could thus better recognise specialised field terms. 
Study information fields also showed wide variations in recall, with Customised EXT significantly outperforming other approaches across most datasets. While almost all methods handled structured metadata like titles, authors, and years effectively, extracting population characteristics proved much harder. These details were often buried in narrative text without standard wording. In real meta-analyses, the population details authors report vary depending on their study's focus—some emphasise demographics, others clinical history—making it challenging to use a single extraction approach or to provide guidance to LLMs in explicit prompt instructions. Customised EXT performed better by collating and drawing together scattered mentions of relevant data points through the manuscripts to build complete study information descriptions, while baseline methods often captured only fragments. For quality assessment, performance differences between methods were less dramatic. Since this information is typically clearly labelled with familiar terms (like "blinded" or "random sequence generation"), even simpler extraction methods achieved relatively high recall and precision. Combined EXT retrieved the most fields in most datasets for this category (MA1, MA2, MA3, and MA6), showing that mixing multiple extraction inputs worked effectively for typical quality assessment content structure in published papers. EXT+Self-reflection provided little extra value. Precision stayed largely stable across all methods and datasets.

\subsubsection{LLM Model Comparison}
\begin{figure}[htbp]
    \centering
    \includegraphics[width=\textwidth]{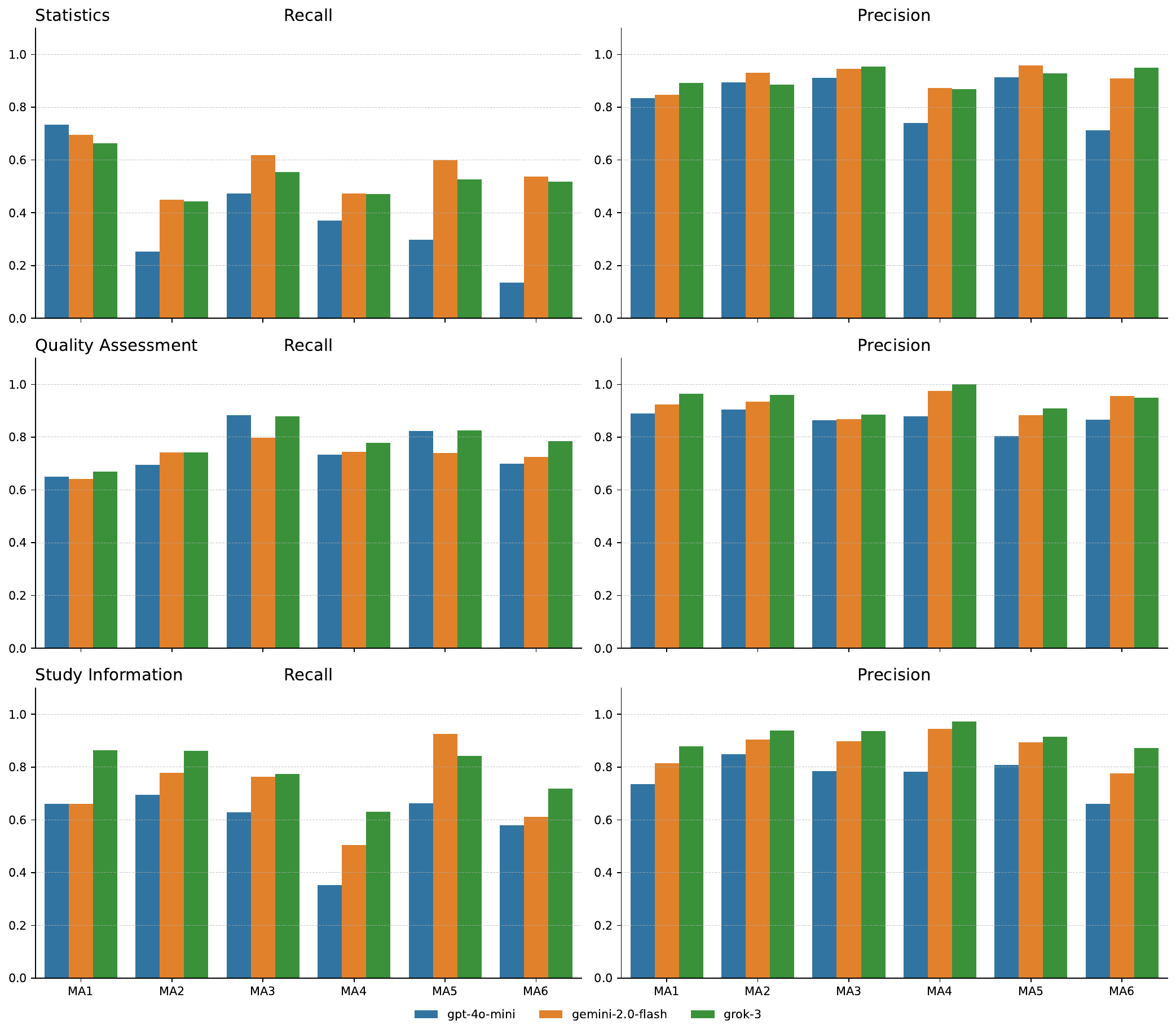}
    \caption{Models Comparison Across Meta-analyses}
    \label{fig:model_per}
\end{figure}

We also assessed the strengths of different models from the three task categories. As shown in Figure \ref{fig:model_per}, the statistics category showed the biggest differences in model performance for recall. Grok and Gemini outperformed GPT, with Gemini achieving the highest recall across most datasets (MA2, MA3, MA4, MA5). These datasets included scattered numerical data, technical terms, and specialised abbreviations (like LDL-C, BTM), which likely made extraction difficult for less capable models like GPT. GPT's recall typically stayed below 0.5, dropping to around 0.16 in MA6. All three models maintained high precision (typically above 0.75), showing that better recall did not lead to extracting irrelevant data. For quality assessment fields, model performance was similar. All models achieved comparable recall and precision across datasets, likely because these fields use standardised formats across studies. Small differences appeared, such as slightly better recall for GPT in MA3 and MA5, and near-perfect precision for Grok in MA4. The study information category also showed notable differences in model performance, particularly for recall. Grok performed best in most datasets except MA5. Gemini also extracted fields effectively across datasets, achieving the highest recall of 0.96 in MA5, while GPT performed less well. Precision remained stable across models, though GPT showed slightly lower precision compared to Gemini and Grok across all datasets.

\subsubsection{Comparison of Method–Model Combinations}
We analysed how the most effective model-method combinations were affected by data category, identifying the configuration with the highest recall for each meta-analysis (Table \ref{tab:best_combos}). In the statistics category, Gemini paired with Customised EXT achieved the highest recall in four of six datasets (MA2, MA3, MA5, MA6), while Grok led in MA4, and GPT performed best in MA1. This shows that for complex statistical data, the model's capabilities matter more than the prompting method, even against advanced data extraction approaches. For quality assessment, Grok was the most frequent top performer, especially when paired with either Customised EXT or EXT. Notably, the best result in MA3 came from Combined EXT using model aggregation, suggesting that blending outputs from multiple models can help when field boundaries are not clearly defined. Precision remained high across all setups (often above 0.95), reflecting the more predictable structure of these fields. For study information, Customised EXT delivered the best performance in most datasets except MA4, indicating its strength in handling descriptive and fragmented fields like population characteristics or eligibility criteria. However, the best model varied by dataset: Grok excelled in MA1, MA4, and MA6, while Gemini led in MA2, MA3, and MA5. The absence of GPT among top performers in this category may suggest that older models struggle with study details, where understanding context and combining information from multiple sentences is essential.

{
\begin{table}[ht]
\centering
\caption{Best-performing model–method combinations by category}
\label{tab:best_combos}
\fontsize{8pt}{10pt}\selectfont
\begin{tabular}{lcccc}
\toprule
\textbf{Dataset} & \textbf{Method} & \textbf{Model} & \textbf{Precision} & \textbf{Recall} \\
\hline\hline
\multicolumn{5}{l}{\textit{Statistics Results}} \\
\hline\hline
MA1 & Customised EXT & GPT        & 0.840 & 0.792 \\
MA2 & Customised EXT & Gemini   & 0.947 & 0.608 \\
MA3 & Customised EXT & Gemini   & 0.927 & 0.893 \\
MA4 & Customised EXT & Grok             & 0.970 & 0.674 \\
MA5 & Customised EXT & Gemini   & 0.986 & 0.799 \\
MA6 & Customised EXT & Gemini   & 0.972 & 0.785 \\
\hline\hline
\multicolumn{5}{l}{\textit{Quality Assessment Results}} \\
\hline\hline
MA1 & Customised EXT & Grok             & 1.000 & 0.684 \\
MA2 & Customised EXT & Grok             & 0.962 & 0.773 \\
MA3 & Combined EXT   & Combined           & 0.871 & 0.900 \\
MA4 & EXT            & Grok             & 1.000 & 0.778 \\
MA5 & Customised EXT & Grok             & 0.935 & 0.878 \\
MA6 & EXT            & Grok             & 0.955 & 0.810 \\
\hline\hline
\multicolumn{5}{l}{\textit{Study Information Results}} \\
\hline\hline
MA1 & Customised EXT & Grok             & 0.826 & 0.919 \\
MA2 & Customised EXT & Gemini   & 0.841 & 0.894 \\
MA3 & Customised EXT & Gemini   & 0.903 & 0.829 \\
MA4 & EXT+Self-reflection & Grok        & 0.980 & 0.639 \\
MA5 & Customised EXT & Gemini   & 0.926 & 0.965 \\
MA6 & Customised EXT & Grok             & 0.842 & 0.882 \\
\bottomrule
\end{tabular}
\end{table}

}

\subsection{Error Distribution}
While overall performance metrics like recall and precision show how well each method works, they do not fully reveal where and what types of errors occur. To better understand the limitations of different data extraction approaches, we investigated  how different types of errors were distributed across specific fields and models. This section breaks down the error distribution with a focus on general trends. Additional analyses, including field-specific differences and the interaction between models and methods, are presented in the appendix for further reference.

\subsubsection{Overall Analysis}
We analysed how errors were distributed across all extracted fields. Of all error cases we analysed, the majority—87.8\% (19,470 instances)—were missing fields. This strongly indicates that current methods and LLM combinations are still challenged to reliably find and extract all relevant fields for fully automating meta-analyses, especially when information is buried in narrative text, spread across multiple sentences and paragraphs, or expressed in non-standard ways. The next most common error was incorrect values, comprising 10.3\% (2,296 instances). These typically involved wrong numerical data, misaligned subgroup details, or misinterpreted statistical results (e.g., reporting the control group sample size as 32 instead of the correct value of 17, reporting 10\% male in total instead of 5\% in the intervention group, or listing separate BMD values instead of the correct combined mean). Less common were overgeneralised errors (e.g., omitting “living independently” from inclusion criteria, or replacing a detailed randomization method such as “1:1 computer-generated list” with the generic term “randomized”) at 1.2\% (267 instances) and incorrect unit errors at 0.7\% (163 instances). From this overall error distribution, we can see that detecting fields (recall) remains the main challenge in current extraction processes, while precision issues, though important, account for a smaller portion of errors. For more detailed analysis, please refer to Table \ref{tab:field-errors} and Table \ref{tab:model_errors} in appendix.

\section{Discussion}
This study evaluated how well LLMs perform structured data extraction for AMA, aiming to identify their current capabilities and key weaknesses in practical, cross-domain situations. We summarise the main findings here, elaborate on the implications and present a recommendation guideline based on our results for guiding researchers on how to mitigate risks when using LLMs for the automation of data extraction in meta-analysis.

\subsection{How Reliable are Current LLMs for Automated Structured Data Extraction in MAs?}

Current state-of-the-art commercial LLMs offer only partial reliability for structured data extraction in meta-analyses. While they show consistent performance on simpler study-level characteristics and some risk-of-bias items, their extraction of statistical results remains error-prone and incomplete, making them unsuitable as standalone tools for end-to-end evidence synthesis without human intervention. These findings highlight the need to carefully examine where LLMs succeed or struggle across different aspects of data extraction, including common errors, data type differences, prompting strategies, model variation, and implications for practical use.

\subsubsection*{What are the most common data extraction errors?}
The frontier commercial LLM models we investigated in this study, most commonly made errors of omission rather than commission. While precision remained high across different LLMs, recall suffered due to missing or partially extracted fields, especially when dispersed or embedded in complex text. Incorrect values were the next major issue, often reflecting misinterpretation of group-specific statistics. Overgeneralisation and unit errors were rare but indicate challenges in preserving specificity and format. Overall, recall limitations remain the primary obstacle to reliable automated data extraction.

\subsubsection*{How reliable are LLMs at extracting different types of data categories?}
Extraction reliability depended on the data categories. Structured, numerical fields such as statistical outcomes were easier for models like Gemini to extract. In contrast, quality assessments and study characteristics required more contextual reasoning, where Grok performed better. Grok achieved the highest recall and precision in those areas. GPT consistently lagged behind in all three categories. Its data extraction performance led to fewer incorrect values but also much lower recall, especially in complex fields. These differences highlighted that no single model performed equally well across all meta-analytic tasks. Each LLM had unique strengths and is therefore better suited to specific types of information or evidence-synthesis tasks.

\subsubsection*{To what extent do different prompting strategies affect performance?}
Prompting strategies significantly influenced extraction quality. Self-reflection led to small gains in recall, usually only 1-2 percentage points. It offered a possible way to recover hard-to-detect content with minimal human involvement, but to work well, it must be narrowly focused. It was more helpful for refining extractions in ambiguous or subjective fields like risk of bias, but had little effect when the original output was already strong. Combining outputs from different models gave more reliable improvements, raising recall by about 5\% on average. Combining their outputs helped build a more complete picture, especially in cases where no single model captured everything on its own. Meta-analysis needed both completeness and reliability, this strategy offered a practical fallback. Model disagreement helped identify uncertain cases for review, while agreement strengthened confidence in the result. Rather than relying on a single system, combining outputs provided a flexible way to improve performance without requiring complex fine-tuning. Domain-specific prompts had the strongest effect. By focusing the model on important fields (e.g. different medical domains, baseline characteristics, and trial outcomes), recall improved, especially in more complex categories like statistical outcomes. However, in some cases this led to a slight drop in precision, as models included broader or less certain information. This trade-off is common in extraction tasks \cite{Grothey2025,Ibrahim2024harnessing} and should be managed depending on whether completeness or accuracy is more important. Customised prompting offered a low-cost, adaptable way to raise extraction quality without altering the underlying model.

\subsubsection*{How variable are the performances of different cutting-edge LLMs?}
GPT tended to extract less data but generated relatively few mistakes. Grok focused on accuracy and achieved the highest precision in statistical extraction (0.939), though with limited recall. Gemini, by contrast, offered a more balanced outcome. It accepted some uncertainty in exchange for broader extraction, resulting in moderate recall while keeping precision within a reasonable range. This approach produced a more complete dataset at the cost of a slight drop in accuracy, which is an acceptable trade-off when a full evidence capture is required. In some ways, Gemini exhibited patterns akin to a human reviewer, scanning widely and using context to pick up subtle clues throughout the paper.

\subsubsection*{What does this mean for practice?}
Customised prompting emerged as the most promising strategy for improving structured data extraction with LLMs. By matching prompts with highlight fields most relevant to meta-analytic synthesis, such as statistical outcomes, LLMs were better guided towards useful outputs. This approach consistently improved recall with only tiny drop in precision, making it particularly valuable for tasks where completeness is essential. Used thoughtfully, task-specific prompts moved LLM-based extraction closer to real-world applications, helping bridge the gap between promising model performance and the demands of AMA. In addition, model selection should align with the nature of the task \cite{wang-etal-2025-match, FernandezPichel2025}. Gemini might by more effective for capturing broader statistical data, Grok was better suited for for analysing quality assessment or reviewing protocols. GPT, while more limited, might still be useful in fast, low-risk screenings or as part of a larger system that includes human oversight. Combining outputs from multiple or integrating reflection step, further enhanced performance in some cases, especially for weaker base models. However, even with these improvements, human oversight remains necessary, which could help build workflows that were both fast and trustworthy, especially for large-scale or time-sensitive reviews. 

\subsection{What Level of Automation Is Enough? Recommendations for Meta-Analytic Extraction}  \label{subsec:task-specific}
\begin{figure}[htbp]
    \centering
    \includegraphics[width=0.8\textwidth]{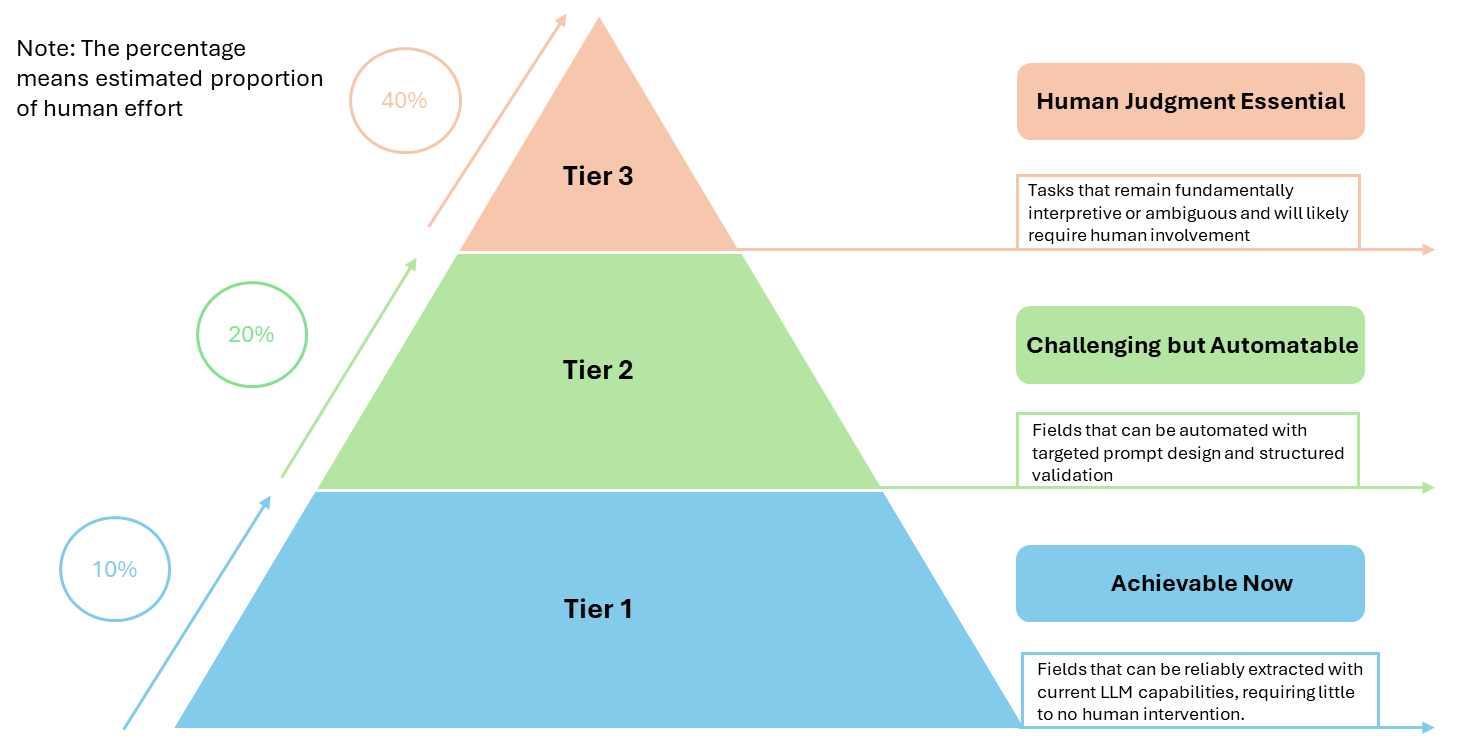}
    \caption{Three-tier automation guideline for structured data extraction in meta-analysis, based on task difficulty, error risk, and need for human oversight. Percentages on the left indicate estimated proportions of total human effort required for each tier for verification of extracted data.}
    \label{fig:tiers}
\end{figure}

The concept of fully automated meta-analytic extraction imagines a system capable of capturing every relevant data point with near-perfect accuracy. But in practice, not all fields in a meta-analysis carry equal weight or require the same level of reliability to support valid conclusions. Some elements, such as statistical results, directly shape pooled estimates and clinical inferences, while others, like study context or author details, primarily serve descriptive or organisational purposes. Errors in the former category have a high risk to the integrity of meta-analytic conclusions, while inaccuracies in the latter may have limited impact. Therefore, in conjunction with our findings, we propose structured, task-specific standards according to the different level of automation in meta-analytic data extraction. These suggestions are not only technical, they reflect how real-world meta-analyses are read and used. Rather than aiming for perfection, extraction for AMA could be based on task-matched performance. We organise information types into three tiers, based on their suitability for automation, the extent of human judgement required and the risk associated with errors. As shown in Figure \ref{fig:tiers}, these tiers represent a spectrum from routine, high-precision extraction (Tier 1), to tasks necessitating targeted oversight (Tier 2), to areas where interpretive reasoning is essential with maximal human-in-the-loop involvement (Tier 3).

\textbf{Tier 1 – Achievable Now} includes fields that are often clearly reported, structurally straightforward, and tolerant of minor errors. Examples include \textit{study information} such as population characteristics, trial location, and author details. These elements mainly aid filtering and contextual understanding.  Because the risk of downstream impact from inaccuracies in these fields is low, our evaluations indicated that even basic prompting yields dependable performance (recall 72–85\%, precision 78–98\%), and reviewer effort could thus be limited to glancing checks of flagged items. These fields are well-suited for efficient, high-volume automation.   

\textbf{Tier 2 – Challenging but Automatable} includes tasks such as \textit{quality assessment}. These involve drawing inferences, understanding intent, and synthesising incomplete information, which can be helped with semi-structured frameworks, such as RoB 2 \cite{Sterne2019RoB2} or GRADE \cite{Guyatt2011GRADE}. Errors in this tier carry moderate risk, especially if misjudgments affect inclusion criteria or bias assessments. LLMs can assist by retrieving relevant rationales and pre-filling fields. In this tier, human review is not optional but essential for accurate interpretation. Additionally, we found that LLMs, especially when guided by combined or reflective prompts, could help bring up relevant content. Using LLMs to pre-fill fields, point out supporting evidence, or mark uncertain judgments can be one way they significantly ease the workload for reviewers. 

\textbf{Tier 3 – Human Judgment Essential} includes fields central to meta-analytic synthesis, such as \textit{statistical results} like effect sizes, confidence intervals, and heterogeneity measures. These demand high recall to capture all outcomes and high precision to avoid distorting effect estimates. Our experiments indicated that customised prompting could achieve moderate performance (recall approximately 76\%, precision ranging from 81\% to 92\%), but even minor extraction errors in this domain could lead to substantial downstream bias. This tier involves the highest risk, as inaccuracies can directly compromise the validity of pooled analyses and clinical conclusions. As a result, tasks in this tier demand structured output formats, clear linkage to source text for verification, and mandatory human verification on primary effect estimates. Automation is feasible here, but only as part of a human-in-the-loop pipeline where reviewers retain responsibility for the final confirmation of each data point.

Table \ref{tab:automation_roadmap} summarises these tiers, linking each information category to its role in meta-analysis, automation goals, reviewer strategy, and roadmap classification. This framework supports a modular AMA development path: starting with Tier 1 fields where performance is already effective, refining Tier 2 domains through task-specific improvements, and supporting Tier 3 tasks through assistive human-in-the-loop workflows. In this view, these results show that future development for AMA is best based on modular extraction process, in which LLM outputs are directed to different post-processing steps depending on field type and confidence level. Progress toward AMA is less about reaching perfection everywhere, and more about knowing where “good enough” might be enough. This approach will enable AMA to advance step by step, beginning with categories that are already achievable, while refining others through focused model adjustments and task-specific training.

{
\begin{table}[htbp]
\centering
%\small
\fontsize{8pt}{10pt}\selectfont
\caption{Automation priorities, strategies, and roadmap tiers by information category}
\begin{tabular}{p{1.5cm}p{3cm}p{6.2cm}p{2.5cm}}
\toprule
\textbf{Category} & \textbf{Role \& Sensitivity} & \textbf{Automation Plan (Target + Reviewer Strategy)} & \textbf{Roadmap Tier} \\
\midrule
Statistical & Core to meta-analytic conclusions; errors distort effec t estimates; low tolerance for mistakes & Recall $\geq$ 90\%, Precision $\geq$ 95\%. \newline Use customised prompts to extract data with linked source text. Human reviewers must verify effect direction, completeness, and comparators. & Tier 3: Challenging; requires prompt tuning \& manual validation \\
\addlinespace[0.5ex]
Quality Assessment & Supports bias judgments; interpretive and semi-subjective; moderate-to-low consistency & Recall $\geq$ 85\%, Precision $\geq$ 95\%. \newline Focus on rationale extraction. Use ensemble methods or self-reflection to retrieve supporting evidence. LLM may suggest labels, but reviewer must interpret vague or implicit reporting. Human input essential in unclear cases. & Tier 2: Automation feasible \\
\addlinespace[0.5ex]
Study Information & Contextual metadata; often explicit; high tolerance for small errors & Recall $\geq$ 80\%, Precision $\geq$ 90\%. \newline Use generic prompts or self-reflection to extract locations, population, study details. Allow auto-accept unless low confidence. Review optional. & Tier 1: Achievable now with minimal oversight \\
\bottomrule
\end{tabular}
\label{tab:automation_roadmap}
\end{table}
}

\subsection{Beyond Extraction: Advancing Dependable and Ethical AMA}
With the rapid advancement of LLMs, AMA has emerged as a key area in evidence synthesis. However, AMA extends beyond individual extraction tasks. It functions as a multi-stage system involving interconnected steps, from study identification and screening to knowledge synthesis \cite{li2025transforming}. With well-defined prompts and clear input formats, LLMs have the potential to make extraction become a practical starting point for automation. Risk-of-bias assessment, while more interpretive, can be supported by semi-structured framework, such as RoB 2 \cite{Sterne2019RoB2} or GRADE \cite{Guyatt2011GRADE}, with LLMs assisting in applying formal criteria and reducing reviewer variability. The most complex stage is evidence synthesis, involving selecting appropriate models (e.g., fixed- vs. random-effects), evaluating heterogeneity, performing sensitivity or subgroup analyses, and integrating findings with clinical context. Existing studies provide limited evidence on how LLMs can contribute to meta-analytic decision-making, not because it is unsuitable for automation, but due to unresolved theoretical and practical challenges \cite{li2025transforming}. Recent developments in chain-of-thought and agent-based LLM systems indicate that these models may offer more capabilities for supporting complex synthesis tasks \cite{wu2025agenticreasoningreasoningllms, lu2024tartopensourcetoolaugmentedframework}. Therefore, future efforts should focus on experimental applications like proposing sensitivity analyses and creating traceable analytical records to enhance transparency and reproducibility while allowing human experts to maintain authority over the core of evidence synthesis. The value of full AMA lies not in replacing expert judgment but in developing frameworks that make such judgments more traceable, repeatable, and transparent.

From an ethical perspective, LLMs in AMA raise concerns about how models present evidence. LLMs can produce seemingly authoritative outputs even with uncertain data \cite{kim2025medicalhallucinationsfoundationmodels}, potentially causing users to place undue trust in conclusions. Though recent reasoning-capable LLMs show more observable logic, these steps don't guarantee interpretability or validity \cite{wei-jie-etal-2024-interpretable}, challenging evidence-based practice principles. AMA systems should incorporate transparency features \cite{Luo2024potential}, including uncertainty annotation, audit trails documenting output production, and model versioning for replication. Ethical use requires cautious automation that is open about processes and accountable to methodological standards. 

Regarding methodological accuracy, we implemented careful measures to prevent feedback loops when using Gemini across multiple tasks. In the merging and evaluation stages, Gemini only operated on JSON-structured data without access to original texts or model identity. This design deliberately prevents the model from using prior knowledge of how values were generated. Additionally, our blinded human validation of 900 samples showed 96.1\% agreement with the LLM-assigned labels, reinforcing the dependability of our evaluation framework. We selected Gemini for these roles not for uniformity but because it offered balanced precision–recall performance across information categories, making it an appropriate tool for non-inferential aggregation and scoring tasks.

\subsection{Study Limitations}
This study has several limitations. First, although we evaluated LLMs on full-text RCTs, our dataset was drawn from a limited number of published meta-analyses across three clinical domains. Broader generalizability remains to be tested. Another limitation is that some key study data was embedded in charts, figures or non-machine-readable graphical elements  which presented an additional challenge. Future work will focus on delving deeper into exploring automated data extraction from these sources specifically. The models we tested were treated as black-box systems, without fine-tuning or access to intermediate model outputs. As a result, we could not directly diagnose why specific fields were omitted, or trace certain failure patterns back to model internals. Also, we did not systematically evaluate computational costs, API expenses, or processing times. Future work will also factor in these operational aspects. Newer models are continuously emerging with greater capabilities, and it is likely that the latest models leveraging test-time compute for increased \enquote{thinking} during inference, may achieve better performances than our results. 

\section{Conclusion}
This study evaluated structured data extraction capabilities of three state-of-the-art commercial LLMs for meta-analysis, comparing their performance across different data types and prompting strategies. While all models achieved high precision, recall remained limited due to frequent omission of key information. Crucially, we found that customised prompts significantly improved data extraction quality, highlighting prompt engineering as essential for enhancing LLM effectiveness in real-world meta-analyses. Based on these insights, we developed clear, three-tiered guidelines data extraction in meta-analysis, matching appropriate levels of LLM-based automation to specific tasks according to their complexity and error risk. Our results emphasised that effective automation of data extraction required not only capable models, but also clearly defined prompts and targeted human oversight.

\bibliographystyle{unsrtnat}
\bibliography{references}

\begin{thebibliography}{57}
\providecommand{\natexlab}[1]{#1}
\providecommand{\url}[1]{\texttt{#1}}
\expandafter\ifx\csname urlstyle\endcsname\relax
  \providecommand{\doi}[1]{doi: #1}\else
  \providecommand{\doi}{doi: \begingroup \urlstyle{rm}\Url}\fi

\bibitem[Cooper(2017)]{cooper_research_2017}
Harris Cooper.
\newblock \emph{Research {Synthesis} and {Meta}-{Analysis}: {A} {Step}-by-{Step} {Approach}}.
\newblock SAGE Publications, Inc, Thousand Oaks, California, 5 edition, 2017.
\newblock \doi{10.4135/9781071878644}.
\newblock URL \url{https://methods.sagepub.com/book/mono/research-synthesis-and-meta-analysis-5e/toc}.

\bibitem[Deeks et~al.(2019)Deeks, Higgins, Altman, and on~behalf of~the Cochrane Statistical Methods~Group]{deeks2019meta}
Jonathan~J Deeks, Julian~PT Higgins, Douglas~G Altman, and on~behalf of~the Cochrane Statistical Methods~Group.
\newblock \emph{Analysing data and undertaking meta-analyses}, chapter~10, pages 241--284.
\newblock John Wiley \& Sons, Ltd, 2019.
\newblock ISBN 9781119536604.
\newblock \doi{https://doi.org/10.1002/9781119536604.ch10}.
\newblock URL \url{https://onlinelibrary.wiley.com/doi/abs/10.1002/9781119536604.ch10}.

\bibitem[Borah et~al.(2017)Borah, Brown, Capers, and Kaiser]{borah_analysis_2017}
Rohit Borah, Andrew~W Brown, Patrice~L Capers, and Kathryn~A Kaiser.
\newblock Analysis of the time and workers needed to conduct systematic reviews of medical interventions using data from the prospero registry.
\newblock \emph{BMJ Open}, 7\penalty0 (2), 2017.
\newblock ISSN 2044-6055.
\newblock \doi{10.1136/bmjopen-2016-012545}.
\newblock URL \url{https://bmjopen.bmj.com/content/7/2/e012545}.

\bibitem[Higgins et~al.(2019)Higgins, Thomas, Chandler, Cumpston, Li, Page, and Welch]{higgins_cochrane_2019}
{Julian P.T.} Higgins, James Thomas, Jacqueline Chandler, Miranda Cumpston, Tianjing Li, {Matthew J.} Page, and {Vivian A.} Welch.
\newblock \emph{Cochrane handbook for systematic reviews of interventions}.
\newblock wiley, January 2019.
\newblock ISBN 9781119536628.
\newblock \doi{10.1002/9781119536604}.
\newblock Publisher Copyright: {\textcopyright} 2019 The Cochrane Collaboration.

\bibitem[Xu et~al.(2022)Xu, Yu, Furuya-Kanamori, Lin, Zorzela, Zhou, Dai, Loke, and Vohra]{xu_validity_data_2022}
Chang Xu, Tianqi Yu, Luis Furuya-Kanamori, Lifeng Lin, Liliane Zorzela, Xiaoqin Zhou, Hanming Dai, Yoon Loke, and Sunita Vohra.
\newblock Validity of data extraction in evidence synthesis practice of adverse events: reproducibility study.
\newblock \emph{BMJ}, 377, 2022.
\newblock \doi{10.1136/bmj-2021-069155}.
\newblock URL \url{https://www.bmj.com/content/377/bmj-2021-069155}.

\bibitem[Marshall and Wallace(2019)]{marshall2019toward}
Iain~J Marshall and Byron~C Wallace.
\newblock Toward systematic review automation: a practical guide to using machine learning tools in research synthesis.
\newblock \emph{Systematic reviews}, 8:\penalty0 1--10, 2019.
\newblock URL \url{https://doi.org/10.1186/s13643-019-1074-9}.

\bibitem[Kiritchenko et~al.(2010)Kiritchenko, De~Bruijn, Carini, Martin, and Sim]{kiritchenko2010exact}
Svetlana Kiritchenko, Berry De~Bruijn, Simona Carini, Joel Martin, and Ida Sim.
\newblock Exact: automatic extraction of clinical trial characteristics from journal publications.
\newblock \emph{BMC medical informatics and decision making}, 10:\penalty0 1--17, 2010.
\newblock \doi{10.1186/1472-6947-10-56}.

\bibitem[Summerscales et~al.(2011)Summerscales, Argamon, Bai, Hupert, and Schwartz]{summerscales2011automatic}
Rodney~L. Summerscales, Shlomo Argamon, Shangda Bai, Jordan Hupert, and Alan Schwartz.
\newblock Automatic summarization of results from clinical trials.
\newblock In \emph{2011 IEEE International Conference on Bioinformatics and Biomedicine}, pages 372--377, 2011.
\newblock \doi{10.1109/BIBM.2011.72}.

\bibitem[Boyko et~al.(2016)Boyko, Kaidina, Kim, Lupatov, Panov, Suvorov, and Shvets]{boyko_framework_2016}
A.~A. Boyko, A.~M. Kaidina, Y.~C. Kim, A.~Yu. Lupatov, A.~I. Panov, R.~E. Suvorov, and A.~V. Shvets.
\newblock A framework for automated meta-analysis: Dendritic cell therapy case study.
\newblock In \emph{2016 IEEE 8th International Conference on Intelligent Systems (IS)}, pages 160--166, 2016.
\newblock \doi{10.1109/IS.2016.7737416}.

\bibitem[Lorenz et~al.(2017)Lorenz, Abdi, Scheckenbach, Pflug, Bülbül, Catapano, Agewall, Ezhov, Bots, Kiechl, Orth, Norata, Empana, Lin, McLachlan, Bokemark, Ronkainen, Amato, Schminke, Srinivasan, Lind, Kato, Dimitriadis, Przewlocki, Okazaki, Stehouwer, Lazarevic, Willeit, Yanez, Steinmetz, Sander, Poppert, Desvarieux, Ikram, Bevc, Staub, Sirtori, Iglseder, Engström, Tripepi, Beloqui, Lee, Friera, Xie, Grigore, Plichart, Su, Robertson, Schmidt, Tuomainen, Veglia, Völzke, Nijpels, Jovanovic, Willeit, Sacco, Franco, Hojs, Uthoff, Hedblad, Park, Suarez, Zhao, Ducimetiere, Chien, Price, Bergström, Kauhanen, Tremoli, Dörr, Berenson, Papagianni, Kablak-Ziembicka, Kitagawa, Dekker, Stolic, Polak, Sitzer, Bickel, Rundek, Hofman, Ekart, Frauchiger, Castelnuovo, Rosvall, Zoccali, Landecho, Bae, Gabriel, Liu, Baldassarre, and Kavousi]{lorenz_automatic_2017}
M.~W. Lorenz, N.~A. Abdi, F.~Scheckenbach, A.~Pflug, A.~Bülbül, A.~L. Catapano, S.~Agewall, M.~Ezhov, M.~L. Bots, S.~Kiechl, A.~Orth, G.~D. Norata, J.~P. Empana, H.~J. Lin, S.~McLachlan, L.~Bokemark, K.~Ronkainen, M.~Amato, U.~Schminke, S.~R. Srinivasan, L.~Lind, A.~Kato, C.~Dimitriadis, T.~Przewlocki, S.~Okazaki, C.~D.~A. Stehouwer, T.~Lazarevic, P.~Willeit, D.~N. Yanez, H.~Steinmetz, D.~Sander, H.~Poppert, M.~Desvarieux, M.~A. Ikram, S.~Bevc, D.~Staub, C.~R. Sirtori, B.~Iglseder, G.~Engström, G.~Tripepi, O.~Beloqui, M.~S. Lee, A.~Friera, W.~Xie, L.~Grigore, M.~Plichart, T.~C. Su, C.~Robertson, C.~Schmidt, T.~P. Tuomainen, F.~Veglia, H.~Völzke, G.~Nijpels, A.~Jovanovic, J.~Willeit, R.~L. Sacco, O.~H. Franco, R.~Hojs, H.~Uthoff, B.~Hedblad, H.~W. Park, C.~Suarez, D.~Zhao, P.~Ducimetiere, K.~L. Chien, J.~F. Price, G.~Bergström, J.~Kauhanen, E.~Tremoli, M.~Dörr, G.~Berenson, A.~Papagianni, A.~Kablak-Ziembicka, K.~Kitagawa, J.~M. Dekker, R.~Stolic, J.~F. Polak, M.~Sitzer, H.~Bickel, T.~Rundek, A.~Hofman,
  R.~Ekart, B.~Frauchiger, S.~Castelnuovo, M.~Rosvall, C.~Zoccali, M.~F. Landecho, J.~H. Bae, R.~Gabriel, J.~Liu, D.~Baldassarre, and M.~Kavousi.
\newblock Automatic identification of variables in epidemiological datasets using logic regression.
\newblock \emph{BMC Med Inform Decis Mak}, 17\penalty0 (1), 2017.
\newblock ISSN 14726947 (ISSN).
\newblock \doi{10.1186/s12911-017-0429-1}.

\bibitem[Michelson(2014)]{michelson_automating_2014}
Matthew Michelson.
\newblock Automating meta-analyses of randomized clinical trials: a first look.
\newblock In \emph{2014 {AAAI} {Fall} {Symposium} {Series}}, 2014.

\bibitem[Cheng et~al.(2021)Cheng, Katz-Rogozhnikov, Varshney, and Baldini]{lu_cheng_automated_2021}
Lu~Cheng, Dmitriy~A. Katz-Rogozhnikov, Kush~R. Varshney, and Ioana Baldini.
\newblock Automated meta-analysis in medical research: {A} causal learning perspective.
\newblock In \emph{In {ACM} {Conference} on {Health}, {Inference}, and {Learning}}, April 2021.

\bibitem[Mutinda et~al.(2022)Mutinda, Liew, Yada, Wakamiya, and Aramaki]{mutinda_automatic_2022}
F.~W. Mutinda, K.~Liew, S.~Yada, S.~Wakamiya, and E.~Aramaki.
\newblock Automatic data extraction to support meta-analysis statistical analysis: a case study on breast cancer.
\newblock \emph{BMC Med Inform Decis Mak}, 22\penalty0 (1):\penalty0 158, June 2022.
\newblock ISSN 1472-6947.
\newblock \doi{10.1186/s12911-022-01897-4}.

\bibitem[Schmidt et~al.(2023)Schmidt, Finnerty~Mutlu, Elmore, Olorisade, Thomas, and Higgins]{Schmidt2023dataextraction}
L~Schmidt, AN~Finnerty~Mutlu, R~Elmore, BK~Olorisade, J~Thomas, and JPT Higgins.
\newblock Data extraction methods for systematic review (semi)automation: Update of a living systematic review [version 2; peer review: 3 approved].
\newblock \emph{F1000Research}, 10\penalty0 (401), 2023.
\newblock \doi{10.12688/f1000research.51117.2}.

\bibitem[Marshall et~al.(2017)Marshall, Kuiper, Banner, and Wallace]{marshall-etal-2017-automating}
Iain Marshall, Jo{\"e}l Kuiper, Edward Banner, and Byron~C. Wallace.
\newblock Automating biomedical evidence synthesis: {R}obot{R}eviewer.
\newblock In Mohit Bansal and Heng Ji, editors, \emph{Proceedings of {ACL} 2017, System Demonstrations}, pages 7--12, Vancouver, Canada, July 2017. Association for Computational Linguistics.
\newblock URL \url{https://aclanthology.org/P17-4002/}.

\bibitem[{Nested Knowledge}(2023)]{nestedknowledge2023screening}
{Nested Knowledge}.
\newblock The data is in: Deciding when to automate screening in your slr, November 2023.
\newblock URL \url{https://about.nested-knowledge.com/2023/11/10/the-data-is-in-deciding-when-to-automate-screening-in-your-slr/}.
\newblock Accessed: 2025-06-09.

\bibitem[Holub et~al.(2021)Holub, Hardy, and Kallmes]{holub2021toward}
Karl Holub, Nicole Hardy, and Kevin Kallmes.
\newblock Toward automated data extraction according to tabular data structure: Cross-sectional pilot survey of the comparative clinical literature.
\newblock \emph{JMIR Form Res}, 5\penalty0 (11):\penalty0 e33124, Nov 2021.
\newblock ISSN 2561-326X.
\newblock \doi{10.2196/33124}.

\bibitem[Wang and Luo(2024)]{wang_metamate_2024}
Xue Wang and Gaoxiang Luo.
\newblock {MetaMate}: {Large} {Language} {Model} to the {Rescue} of {Automated} {Data} {Extraction} for {Educational} {Systematic} {Reviews} and {Meta}-analyses, 2024.

\bibitem[OpenAI(2023)]{openai_chatgpt}
OpenAI.
\newblock Chatgpt: Large language model (mar 14 version).
\newblock \url{https://chat.openai.com/}, 2023.
\newblock Accessed May 15, 2025.

\bibitem[Anthropic(2023)]{anthropic2023claude2}
Anthropic.
\newblock Claude 2 model announcement.
\newblock \url{https://www.anthropic.com/news/claude-2}, 2023.
\newblock Accessed: 2025-05-28.

\bibitem[Kartchner et~al.(2023)Kartchner, Ramalingam, Al-Hussaini, Kronick, and Mitchell]{kartchner_zero-shot_2023}
David Kartchner, Selvi Ramalingam, Irfan Al-Hussaini, Olivia Kronick, and Cassie Mitchell.
\newblock Zero-shot information extraction for clinical meta-analysis using large language models.
\newblock In Dina Demner-fushman, Sophia Ananiadou, and Kevin Cohen, editors, \emph{The 22nd Workshop on Biomedical Natural Language Processing and BioNLP Shared Tasks}, pages 396--405, Toronto, Canada, July 2023. Association for Computational Linguistics.
\newblock \doi{10.18653/v1/2023.bionlp-1.37}.
\newblock URL \url{https://aclanthology.org/2023.bionlp-1.37/}.

\bibitem[Konet et~al.(2024)Konet, Thomas, Gartlehner, Kahwati, Hilscher, Kugley, Crotty, Viswanathan, and Chew]{konet2024performance}
Amanda Konet, Ian Thomas, Gerald Gartlehner, Leila Kahwati, Rainer Hilscher, Shannon Kugley, Karen Crotty, Meera Viswanathan, and Robert Chew.
\newblock Performance of two large language models for data extraction in evidence synthesis.
\newblock \emph{Research synthesis methods}, 15\penalty0 (5):\penalty0 818--824, 2024.

\bibitem[Yun et~al.(2024)Yun, Pogrebitskiy, Marshall, and Wallace]{yun_automatically_2024}
Hye~Sun Yun, David Pogrebitskiy, Iain~James Marshall, and Byron~C Wallace.
\newblock Automatically extracting numerical results from randomized controlled trials with large language models.
\newblock In Kaivalya Deshpande, Madalina Fiterau, Shalmali Joshi, Zachary Lipton, Rajesh Ranganath, and Iñigo Urteaga, editors, \emph{Proceedings of the 9th Machine Learning for Healthcare Conference}, volume 252 of \emph{Proceedings of Machine Learning Research}. PMLR, 8 2024.
\newblock URL \url{https://proceedings.mlr.press/v252/yun24a.html}.

\bibitem[Schmidt et~al.(2024)Schmidt, Hair, Graziozi, Campbell, Kapp, Khanteymoori, Craig, Engelbert, and Thomas]{schmidt2025exploringuselargelanguage}
Lena Schmidt, Kaitlyn Hair, Sergio Graziozi, Fiona Campbell, Claudia Kapp, Alireza Khanteymoori, Dawn Craig, Mark Engelbert, and James Thomas.
\newblock Exploring the use of a large language model for data extraction in systematic reviews: a rapid feasibility study.
\newblock In \emph{Proceedings of the 3rd Workshop on Augmented Intelligence for Technology-Assisted Reviews Systems (ALTARS 2024)}, volume 3832, Glasgow, UK, 2024. CEUR-WS.org.
\newblock URL \url{https://ceur-ws.org/Vol-3832/paper2.pdf}.

\bibitem[Matsumoto et~al.(2024)Matsumoto, Nakai, Sakakibara, Aimiya, Sugiura, Lee, Yamada, and Mizuno]{matsumoto_remote_2024}
Noriaki Matsumoto, Tsuyoshi Nakai, Mikio Sakakibara, Yukinori Aimiya, Shinya Sugiura, Jeannie~K. Lee, Shigeki Yamada, and Tomohiro Mizuno.
\newblock Remote follow-up by pharmacists for blood pressure control in patients with hypertension: a systematic review and a meta-analysis of randomized controlled trials.
\newblock \emph{Scientific reports}, 14\penalty0 (1):\penalty0 2535, January 2024.
\newblock ISSN 2045-2322.
\newblock \doi{10.1038/s41598-024-52894-8}.

\bibitem[Guo et~al.(2021)Guo, Li, Yang, Liao, Zhang, Zhu, Zhao, Chen, Zhang, and Lei]{guo_effects_2021}
Ru~Guo, Nian Li, Rong Yang, Xiao-Yang Liao, Yu~Zhang, Ben-Fu Zhu, Qian Zhao, Lingmin Chen, Yong-Gang Zhang, and Yi~Lei.
\newblock Effects of the {Modified} {DASH} {Diet} on {Adults} {With} {Elevated} {Blood} {Pressure} or {Hypertension}: {A} {Systematic} {Review} and {Meta}-{Analysis}.
\newblock \emph{Frontiers in nutrition}, 8:\penalty0 725020, 2021.
\newblock ISSN 2296-861X.
\newblock \doi{10.3389/fnut.2021.725020}.

\bibitem[Khalid et~al.(2023)Khalid, Abdelrahim, Hanach, AlKurd, Khan, Mahrous, Radwan, Naja, Madkour, Obaideen, Khraiwesh, and Faris]{khalid_effect_2023}
Narmin Khalid, Dana~N. Abdelrahim, Nivine Hanach, Refat AlKurd, Moien Khan, Lana Mahrous, Hadia Radwan, Farah Naja, Mohamed Madkour, Khaled Obaideen, Husam Khraiwesh, and MoezAlIslam Faris.
\newblock Effect of camel milk on lipid profile among patients with diabetes: a systematic review, meta-analysis, and meta-regression of randomized controlled trials.
\newblock \emph{BMC Complementary Medicine and Therapies}, 23\penalty0 (1):\penalty0 438, December 2023.
\newblock ISSN 2662-7671.
\newblock \doi{10.1186/s12906-023-04257-5}.
\newblock URL \url{https://bmccomplementalternmed.biomedcentral.com/articles/10.1186/s12906-023-04257-5}.

\bibitem[Yu et~al.(2025)Yu, Fu, Chen, Fang, and Tsai]{yu_effect_2025}
Yu-Ting Yu, Yu-Hsiang Fu, Yi-Hsien Chen, Yu-Wei Fang, and Ming-Hsien Tsai.
\newblock Effect of dietary glycemic index on insulin resistance in adults without diabetes mellitus: a systematic review and meta-analysis.
\newblock \emph{Frontiers in Nutrition}, 12:\penalty0 1458353, February 2025.
\newblock ISSN 2296-861X.
\newblock \doi{10.3389/fnut.2025.1458353}.
\newblock URL \url{https://www.frontiersin.org/articles/10.3389/fnut.2025.1458353/full}.

\bibitem[Kim et~al.(2024)Kim, Choi, Gu, Ko, Kwon, Han, Kim, and Kim]{kim_effects_2024}
Hee-Ju Kim, Seo-A. Choi, Min-Sun Gu, Seo-Yeong Ko, Jae-Hee Kwon, Ja-Young Han, Jae~Hyun Kim, and Myeong~Gyu Kim.
\newblock Effects of {Glucagon}-{Like} {Peptide}-1 {Receptor} {Agonist} on {Bone} {Mineral} {Density} and {Bone} {Turnover} {Markers}: {A} {Meta}-{Analysis}.
\newblock \emph{Diabetes/metabolism research and reviews}, 40\penalty0 (6):\penalty0 e3843, September 2024.
\newblock ISSN 1520-7560 1520-7552.
\newblock \doi{10.1002/dmrr.3843}.

\bibitem[Oldrini et~al.(2022)Oldrini, Feltri, Albanese, Lucchina, Filardo, and Candrian]{oldrini_volar_2022}
Lorenzo~Massimo Oldrini, Pietro Feltri, Jacopo Albanese, Stefano Lucchina, Giuseppe Filardo, and Christian Candrian.
\newblock Volar locking plate vs cast immobilization for distal radius fractures: a systematic review and meta-analysis.
\newblock \emph{EFORT Open Reviews}, 7\penalty0 (9):\penalty0 644--652, September 2022.
\newblock ISSN 2058-5241.
\newblock \doi{10.1530/EOR-22-0022}.
\newblock URL \url{https://eor.bioscientifica.com/view/journals/eor/7/9/EOR-22-0022.xml}.

\bibitem[{OpenAI}(2024)]{openai2024gpt4omini}
{OpenAI}.
\newblock Gpt-4o mini: Advancing cost-efficient intelligence.
\newblock \url{https://openai.com/index/gpt-4o-mini-advancing-cost-efficient-intelligence/}, 2024.

\bibitem[{Google DeepMind}(2024)]{google2024gemini2flash}
{Google DeepMind}.
\newblock Gemini 2.0 flash.
\newblock \url{https://deepmind.google/technologies/gemini/flash/}, 2024.

\bibitem[xAI(2024)]{xai2024grok}
xAI.
\newblock Grok-3 language model.
\newblock \url{https://x.ai/}, 2024.

\bibitem[Windisch et~al.(2024)Windisch, Dennst{\"a}dt, Koechli, Schr{\"o}der, Aebersold, F{\"o}rster, Zwahlen, and Windisch]{windisch2024impact}
Paul Windisch, Fabio Dennst{\"a}dt, Carole Koechli, Christina Schr{\"o}der, Daniel~M Aebersold, Robert F{\"o}rster, Daniel~R Zwahlen, and Paul~Y Windisch.
\newblock The impact of temperature on extracting information from clinical trial publications using large language models.
\newblock \emph{Cureus}, 16\penalty0 (12), 2024.

\bibitem[Schroeder et~al.(2025)Schroeder, Jaldi, and Zhang]{schroeder2025llmloop}
Noah~L. Schroeder, Chris~Davis Jaldi, and Shan Zhang.
\newblock Large language models with human-in-the-loop validation for systematic review data extraction, 2025.
\newblock URL \url{https://arxiv.org/abs/2501.11840}.

\bibitem[Cloud(2025)]{googlecloud2025geminidata}
Google Cloud.
\newblock Use gemini 2.0 to speed up data processing.
\newblock \url{https://cloud.google.com/blog/products/ai-machine-learning/use-gemini-2-0-to-speed-up-data-processing}, 2025.
\newblock Blog post.

\bibitem[Zhang et~al.(2025)Zhang, Wang, and Liu]{zhang2025grokqigong}
Hao Zhang, Lin Wang, and Shiyu Liu.
\newblock Harnessing ai for integrative medicine: Exploring grok 3’s role in researching qigong, tai chi, yoga, and mindfulness for college students’ mental health.
\newblock \emph{American Journal of Biomedical Science and Research}, 26\penalty0 (3), 2025.
\newblock \doi{10.34297/AJBSR.2025.26.003468}.
\newblock URL \url{https://biomedgrid.com/pdf/AJBSR.MS.ID.003468.pdf}.

\bibitem[Microsoft(2025)]{microsoft2025grokazure}
Microsoft.
\newblock Microsoft adds elon musk’s grok-3 to azure, citing healthcare and science use cases.
\newblock \url{https://www.mobihealthnews.com/news/microsoft-adds-elon-musks-grok-3-azure-citing-healthcare-and-science-use-cases}, 2025.
\newblock News release.

\bibitem[Shinn et~al.(2023)Shinn, Cassano, Gopinath, Narasimhan, and Yao]{shinn2023selfreflection}
Noah Shinn, Federico Cassano, Ashwin Gopinath, Karthik Narasimhan, and Shunyu Yao.
\newblock Reflexion: language agents with verbal reinforcement learning.
\newblock In A.~Oh, T.~Naumann, A.~Globerson, K.~Saenko, M.~Hardt, and S.~Levine, editors, \emph{Advances in Neural Information Processing Systems}, volume~36, pages 8634--8652. Curran Associates, Inc., 2023.
\newblock URL \url{https://proceedings.neurips.cc/paper_files/paper/2023/file/1b44b878bb782e6954cd888628510e90-Paper-Conference.pdf}.

\bibitem[Ji et~al.(2023)Ji, Yu, Xu, Lee, Ishii, and Fung]{ji_etal_2023_towards}
Ziwei Ji, Tiezheng Yu, Yan Xu, Nayeon Lee, Etsuko Ishii, and Pascale Fung.
\newblock Towards mitigating {LLM} hallucination via self reflection.
\newblock In Houda Bouamor, Juan Pino, and Kalika Bali, editors, \emph{Findings of the Association for Computational Linguistics: EMNLP 2023}, pages 1827--1843, Singapore, December 2023. Association for Computational Linguistics.
\newblock \doi{10.18653/v1/2023.findings-emnlp.123}.
\newblock URL \url{https://aclanthology.org/2023.findings-emnlp.123/}.

\bibitem[Li et~al.(2024{\natexlab{a}})Li, Yang, and Ettinger]{li_etal_2024_hindsight}
Yanhong Li, Chenghao Yang, and Allyson Ettinger.
\newblock When hindsight is not 20/20: Testing limits on reflective thinking in large language models.
\newblock In Kevin Duh, Helena Gomez, and Steven Bethard, editors, \emph{Findings of the Association for Computational Linguistics: NAACL 2024}, pages 3741--3753, Mexico City, Mexico, June 2024{\natexlab{a}}. Association for Computational Linguistics.
\newblock \doi{10.18653/v1/2024.findings-naacl.237}.
\newblock URL \url{https://aclanthology.org/2024.findings-naacl.237/}.

\bibitem[Dietterich(2000)]{Dietterich2000ensemble}
Thomas~G. Dietterich.
\newblock Ensemble methods in machine learning.
\newblock In \emph{Multiple Classifier Systems}, pages 1--15, Berlin, Heidelberg, 2000. Springer Berlin Heidelberg.
\newblock ISBN 978-3-540-45014-6.

\bibitem[Dong et~al.(2020)Dong, Yu, Cao, Shi, and Ma]{dong2020survey}
Xueying Dong, Zhiwen Yu, Wen Cao, Yanchao Shi, and Qiang Ma.
\newblock A survey on ensemble learning.
\newblock \emph{Frontiers of Computer Science}, 14\penalty0 (2):\penalty0 241--258, 2020.
\newblock \doi{10.1007/s11704-019-8208-z}.

\bibitem[Li et~al.(2024{\natexlab{b}})Li, Wei, Huang, Li, Hu, Chuang, He, Das, Keloth, Yang, Diala, Roberts, Tao, Jiang, Zheng, and Xu]{li2024ensemble}
Zhao Li, Qiang Wei, Liang-Chin Huang, Jianfu Li, Yan Hu, Yao-Shun Chuang, Jianping He, Avisha Das, Vipina~Kuttichi Keloth, Yuntao Yang, Chiamaka~S Diala, Kirk~E Roberts, Cui Tao, Xiaoqian Jiang, W~Jim Zheng, and Hua Xu.
\newblock Ensemble pretrained language models to extract biomedical knowledge from literature.
\newblock \emph{Journal of the American Medical Informatics Association}, 31\penalty0 (9):\penalty0 1904--1911, 03 2024{\natexlab{b}}.
\newblock ISSN 1527-974X.
\newblock \doi{10.1093/jamia/ocae061}.
\newblock URL \url{https://doi.org/10.1093/jamia/ocae061}.

\bibitem[Zhang and Chen(2022)]{zhang2022biomedical}
Z.~Zhang and A.L.P. Chen.
\newblock Biomedical named entity recognition with the combined feature attention and fully-shared multi-task learning.
\newblock \emph{BMC Bioinformatics}, 23\penalty0 (1):\penalty0 458, 2022.
\newblock \doi{10.1186/s12859-022-04994-3}.

\bibitem[Grothey et~al.(2025)Grothey, Odenkirchen, Brkic, et~al.]{Grothey2025}
Bernd Grothey, Jonas Odenkirchen, Ana Brkic, et~al.
\newblock Comprehensive testing of large language models for extraction of structured data in pathology.
\newblock \emph{Communications Medicine}, 5:\penalty0 96, 2025.
\newblock \doi{10.1038/s43856-025-00808-8}.
\newblock URL \url{https://doi.org/10.1038/s43856-025-00808-8}.

\bibitem[Ibrahim et~al.(2024)Ibrahim, Dao, and Shah]{Ibrahim2024harnessing}
Ayyub Ibrahim, Huy Dao, and Tarak Shah.
\newblock Innocence discovery lab - harnessing large language models to surface data buried in wrongful conviction case documents.
\newblock \emph{The Wrongful Conviction Law Review}, 5\penalty0 (1):\penalty0 103--126, 2024.
\newblock \doi{https://doi.org/10.29173/wclawr112}.

\bibitem[Wang et~al.(2025)Wang, Chen, Lin, Chen, Han, Sun, Wang, and Zeng]{wang-etal-2025-match}
Tianshu Wang, Xiaoyang Chen, Hongyu Lin, Xuanang Chen, Xianpei Han, Le~Sun, Hao Wang, and Zhenyu Zeng.
\newblock Match, compare, or select? an investigation of large language models for entity matching.
\newblock In Owen Rambow, Leo Wanner, Marianna Apidianaki, Hend Al-Khalifa, Barbara~Di Eugenio, and Steven Schockaert, editors, \emph{Proceedings of the 31st International Conference on Computational Linguistics}, pages 96--109, Abu Dhabi, UAE, January 2025. Association for Computational Linguistics.
\newblock URL \url{https://aclanthology.org/2025.coling-main.8/}.

\bibitem[Fernández-Pichel et~al.(2025)Fernández-Pichel, Pichel, and Losada]{FernandezPichel2025}
María Fernández-Pichel, José~Carlos Pichel, and David~E. Losada.
\newblock Evaluating search engines and large language models for answering health questions.
\newblock \emph{npj Digital Medicine}, 8:\penalty0 153, 2025.
\newblock \doi{10.1038/s41746-025-01546-w}.
\newblock URL \url{https://doi.org/10.1038/s41746-025-01546-w}.

\bibitem[Sterne et~al.(2019)Sterne, Savović, Page, Elbers, Blencowe, Boutron, Cates, Cheng, Corbett, Eldridge, Emberson, Hernán, Hopewell, Hróbjartsson, Junqueira, Jüni, Kirkham, Lasserson, Li, McAleenan, Reeves, Shepperd, Shrier, Stewart, Tilling, White, Whiting, and Higgins]{Sterne2019RoB2}
Jonathan A.~C. Sterne, Jelena Savović, Matthew~J. Page, Roy~G. Elbers, Natalie~S. Blencowe, Isabelle Boutron, Christopher~J. Cates, Hung-Yuan Cheng, Mark~S. Corbett, Sandra~M. Eldridge, Jonathan~R. Emberson, Miguel~A. Hernán, Sally Hopewell, Asbjørn Hróbjartsson, Daniela~R. Junqueira, Peter Jüni, Jamie~J. Kirkham, Toby Lasserson, Tianjing Li, Alexandra McAleenan, Barnaby~C. Reeves, Sasha Shepperd, Ian Shrier, Lesley~A. Stewart, Kate Tilling, Ian~R. White, Penny~F. Whiting, and Julian P.~T. Higgins.
\newblock Rob 2: a revised tool for assessing risk of bias in randomised trials.
\newblock \emph{BMJ}, 366:\penalty0 l4898, 2019.
\newblock \doi{10.1136/bmj.l4898}.

\bibitem[Guyatt et~al.(2011)Guyatt, Oxman, Akl, Kunz, Vist, Brozek, Chen, and Schünemann]{Guyatt2011GRADE}
Gordon~H. Guyatt, Andrew~D. Oxman, Elie~A. Akl, Regina Kunz, Gunn~E. Vist, Jan Brozek, Yaolong Chen, and Holger~J. Schünemann.
\newblock Grade guidelines: 1. introduction—grade evidence profiles and summary of findings tables.
\newblock \emph{Journal of Clinical Epidemiology}, 64\penalty0 (4):\penalty0 383--394, 2011.
\newblock \doi{10.1016/j.jclinepi.2010.04.026}.

\bibitem[Li et~al.(2025)Li, Mathrani, and Susnjak]{li2025transforming}
Lingbo Li, Anuradha Mathrani, and Teo Susnjak.
\newblock Transforming evidence synthesis: A systematic review of the evolution of automated meta-analysis in the age of ai, 2025.
\newblock URL \url{https://arxiv.org/abs/2504.20113}.

\bibitem[Wu et~al.(2025)Wu, Zhu, and Liu]{wu2025agenticreasoningreasoningllms}
Junde Wu, Jiayuan Zhu, and Yuyuan Liu.
\newblock Agentic reasoning: Reasoning llms with tools for the deep research, 2025.
\newblock URL \url{https://arxiv.org/abs/2502.04644}.

\bibitem[Lu et~al.(2025)Lu, Pan, Ma, Nakov, and Kan]{lu2024tartopensourcetoolaugmentedframework}
Xinyuan Lu, Liangming Pan, Yubo Ma, Preslav Nakov, and Min-Yen Kan.
\newblock {TART}: An open-source tool-augmented framework for explainable table-based reasoning.
\newblock In Luis Chiruzzo, Alan Ritter, and Lu~Wang, editors, \emph{Findings of the Association for Computational Linguistics: NAACL 2025}, pages 4323--4339, Albuquerque, New Mexico, April 2025. Association for Computational Linguistics.
\newblock ISBN 979-8-89176-195-7.
\newblock URL \url{https://aclanthology.org/2025.findings-naacl.244/}.

\bibitem[Kim et~al.(2025)Kim, Jeong, Chen, Li, Lu, Alhamoud, Mun, Grau, Jung, Gameiro, Fan, Park, Lin, Yoon, Yoon, Sap, Tsvetkov, Liang, Xu, Liu, McDuff, Lee, Park, Tulebaev, and Breazeal]{kim2025medicalhallucinationsfoundationmodels}
Yubin Kim, Hyewon Jeong, Shen Chen, Shuyue~Stella Li, Mingyu Lu, Kumail Alhamoud, Jimin Mun, Cristina Grau, Minseok Jung, Rodrigo~R Gameiro, Lizhou Fan, Eugene Park, Tristan Lin, Joonsik Yoon, Wonjin Yoon, Maarten Sap, Yulia Tsvetkov, Paul~Pu Liang, Xuhai Xu, Xin Liu, Daniel McDuff, Hyeonhoon Lee, Hae~Won Park, Samir~R Tulebaev, and Cynthia Breazeal.
\newblock Medical hallucination in foundation models and their impact on healthcare.
\newblock \emph{medRxiv}, 2025.
\newblock \doi{10.1101/2025.02.28.25323115}.
\newblock URL \url{https://www.medrxiv.org/content/early/2025/03/03/2025.02.28.25323115}.

\bibitem[Wei~Jie et~al.(2024)Wei~Jie, Satapathy, Goh, and Cambria]{wei-jie-etal-2024-interpretable}
Yeo Wei~Jie, Ranjan Satapathy, Rick Goh, and Erik Cambria.
\newblock How interpretable are reasoning explanations from prompting large language models?
\newblock In Kevin Duh, Helena Gomez, and Steven Bethard, editors, \emph{Findings of the Association for Computational Linguistics: NAACL 2024}, pages 2148--2164, Mexico City, Mexico, June 2024. Association for Computational Linguistics.
\newblock \doi{10.18653/v1/2024.findings-naacl.138}.
\newblock URL \url{https://aclanthology.org/2024.findings-naacl.138/}.

\bibitem[Luo et~al.(2024)Luo, Chen, Zhu, Wang, Wang, Liu, Lyu, Wang, Wang, and Chen]{Luo2024potential}
Xufei Luo, Fengxian Chen, Di~Zhu, Ling Wang, Zijun Wang, Hui Liu, Meng Lyu, Ye~Wang, Qi~Wang, and Yaolong Chen.
\newblock Potential roles of large language models in the production of systematic reviews and meta-analyses.
\newblock \emph{J Med Internet Res}, 26:\penalty0 e56780, 6 2024.
\newblock ISSN 1438-8871.
\newblock \doi{10.2196/56780}.
\newblock URL \url{https://www.jmir.org/2024/1/e56780}.

\end{thebibliography}

\appendix  
\section{Baseline Extraction Prompt}

\begin{mdframed}[backgroundcolor=gray!10]
\fontsize{8pt}{10pt}\selectfont
\begin{verbatim}

You are a world-leading expert in medical literature data extraction. Your task is to extract 
structured research data from RCT PDFs to enable meta-analysis. The extracted data must be 
formatted precisely, ensuring alignment with meta-analysis requirements.

Input: A full-text research PDF document.

Task: Extract study-related information using a structured approach
   - Study Characteristics (SC): General study details (e.g., study setting, funding source, 
   ethical approval).
   - Participant Characteristics (PC): Demographic and clinical information of participants 
   (e.g., age, gender distribution, baseline health conditions).
   - Intervention/Exposure (IE): Treatment, exposure, or intervention details (e.g., type, 
   dosage, frequency, duration of intervention).
   - Comparison/Control (CC): Description of the control/comparison group (e.g., usual care, 
   placebo).
   - Outcome Measures (OM): Extract all reported outcomes with their values, including time 
   points and statistics.
   - Study Design (SD): Methodology details (e.g., randomisation, blinding [single, double, 
   triple?], allocation concealment, study duration). *Pay special attention to aspects of 
   study design relevant to bias assessment.*

Additional required elements:
   - Study Info: First author, publication year, country.
   - Sample Size: Total and per group. 
   - Eligibility Criteria: Inclusion/exclusion criteria.

Data Extraction Rules
1. Data Sources & Priority
   - Extract from all available sections: abstract, methods, results, tables, figures, 
   appendices(if appaliable).
   - IMPORTANT: Under no circumstances should the LLM attempt to calculate any statistical 
   values.
   - *Source Priority:*
     - *Tables* are the preferred source for numerical values.
     - If a discrepancy exists between tables and text, use table data unless an alternative 
     choice is justified in `"justification"`.
     - If conflicting values exist, store them in `"data_conflicts"` and specify sources.
     - If data is *only* found in a figure, attempt to extract the numerical values. Note 
     limitations in `"notes"`.
   - For *every* extracted data point, include a `"source"` field indicating the section of 
   the paper (e.g., "Table 1", "Results section, paragraph 3", "Methods section", "Figure 
   2"). Be as specific as possible.
   - Add a `"confidence"` field with values "High", "Medium", or "Low". Base this assessment 
   on the clarity of the data in the source, consistency across the paper, and the presence 
   of supporting information. For example:
     - "High":  Data is clearly presented in a table, consistent with the text, and supported 
     by confidence intervals or p-values.
     - "Medium": Data is found in the text but may have minor inconsistencies with other 
     sections, or lacks supporting statistical information.
     - "Low":  Data is unclear, potentially ambiguous, only indirectly inferable, or relies 
     on interpreting a figure without explicit values.
2. Outcome Measures (OM) Standardization
   - Extract outcome descriptions, time points, and numerical values for both intervention 
   and control groups.
   - If multiple time points exist, prioritize the *final* follow-up but list all time points 
   in `"other_time_points"`.
   - Report exact statistics: means (SD), medians (IQR), ranges. *Do not compute new 
   statistics.*
   - Use `"needs_transformation": true` for data requiring conversion (e.g., median & IQR to 
   mean & SD). Otherwise, ensure `"needs_transformation": true` is not set.
   - Include confidence intervals and p-values when available for each group at each time 
   point.
   - Use descriptive and consistent names for outcomes (e.g., `outcome_cognitive_function`, 
   `outcome_physical_function`).
   - Mark whether an outcome is *primary* (`"primary_outcome": true`) or *secondary* 
   (`"primary_outcome": false`).
3. Unit Standardization
   - Ensure all extracted data maintains its original unit. Always report the original unit, 
   even if the source paper is consistent about the unit used.
   - ONLY When unit conversions are needed *or if multiple units are reported or implied*, 
   mark `"needs_transformation": true` and provide the `"standardised_unit"` field. Document 
   the conversion factor in `"notes"` if possible.
4. Handling Missing Data
   - **Numerical Data:** If *numerical* data (e.g., means, standard deviations, sample sizes, 
   p_value) is absent from the PDF, ONLY return `"null"`, NOT `"Not reported"` or `"NA "`. If 
   the missing numerical data can be calculated from other reported data (e.g., SD from CI 
   and Mean), note this in `"needs_transformation"` as `true` and add a `"notes"` field 
   explaining the calculation method.
   - **Non-Numerical Data:** If *non-numerical* data (e.g., study design features, 
   descriptions of interventions, blinding methods) is absent or NOT explicit from the PDF, 
   ONLY return `"Not reported"`.  Do NOT attempt to fill in missing non-numerical values.
5. Adverse Events & Dropouts
   - Extract total adverse events, serious events, and dropout numbers.
   - If adverse events are reported for <3 types, extract exact values for each type. For 
   example, extract cardiovascular, gastrointestinal, neurological, psychiatric adverse 
   events, etc. If events are described qualitatively (e.g., "more common in the intervention 
   group"), note this in `"notes"` and assign a `"confidence"` of "Low".
6. Data Conflict Handling
   - If conflicting values exist, store them in `"data_conflicts"`, specifying the sources.
   - Justify which value is used in `"justification"`. Example:
```json
{
  "data_conflicts": {
    "sample_size": {
      "table": 120,
      "results_section": 115
    }
  },
  "justification": "Table 1 provided the most complete data, including confidence intervals."
}
```
- If data conflicts arise, prioritize: 1. Published errata/corrections, 2. Data from tables 
over text, 3. Data from the 'Results' section over the 'Methods' section when reporting 
outcome values, 4. explicit values over values inferred from figures.
7. PDF Processing Status
If the PDF is unreadable, return:
```json
{ "pdf_status": "Unreadable" }
```
Otherwise, return "pdf_status": "Processed".

Output Format:
Structured JSON with the following keys. Adjust keys as needed to fit specific data to ensure 
you can capture all data needed
- `"title"`: Title of the research paper.
- `"first_author"`: First author's name.
- `"publication_year"`: Year of publication.
- `"country"`: Country where the research was conducted.
- `"study_design"`: A dictionary containing methodology details of the study (e.g., 
randomisation, blinding, study_duration).
- `"sample_size"`: A dictionary containing the total sample size and sample size for each 
group (intervention/exposure vs. comparison/control). 
- `"study_characteristics"`: A dictionary containing key study details (e.g. study setting, 
funding source, and ethical approval). 
- `"participant_characteristics"`: A dictionary containing participant-specific details(e.g. 
age, gender distribution, baseline health conditions).
- `"intervention_exposure"`: A dictionary containing details about the intervention or 
exposure being studied, including duration.
- `"comparison_control"`: A dictionary containing details about the control or comparison 
group.
- `"outcome_cognitive_function"`(and similar outcomes using consistent naming):  A dictionary 
that includes all reported outcomes from the study. Extract primary_outcome, final_followup 
and other_time_points where reported.
- `"adverse_events"`: A dictionary that contains the number of total adverse events, serious 
events, and specific types of adverse events reported during the study.
- `"dropouts"`: A dictionary containing the number of participants who dropped out of the 
study.
- `"eligibility_criteria"`: A dictionary containing both the inclusion and exclusion criteria.
- `"pdf_status"`:  "Processed" if the PDF was readable; "Unreadable" otherwise.
- `"notes"`: Any additional or clarifying notes made during data extraction.

Here's an example output:
```json
{
  "title": "The Impact of Regular Exercise on Cognitive Function in Older Adults",
  "first_author": "Jane Doe",
  "publication_year": 2023,
  "country": "USA",
  "study_design": {
      "randomisation": "Randomised assignment to intervention or control group",
      "blinding": "Double-blinded",
      "allocation_concealment": "Adequate",
      "study_duration": "12 months",
      "source": "Methods section, paragraph 2",
      "confidence": "High"
  },
  "sample_size": {
    "total": 100,
    "intervention_group": 50,
    "control_group": 50,
    "source": "Table 1",
    "confidence": "High"
  },
  "study_characteristics": {
    "study_setting": "Community center",
    "funding_source": "National Institutes of Health",
    "ethical_approval": "Institutional Review Board approved",
    "source": "Methods section, paragraph 1",
    "confidence": "High"
  },
  "participant_characteristics": {
    "age": {
        "intervention":{
          "mean": 72.5,
          "sd": 6.2,
          "unit": "years"
        },
        "control":{
          "mean": 73.1,
          "sd": 5.8,
          "unit": "years"
        },
      "source": "Table 1",
      "confidence": "High"
    },
    "gender_distribution": {
      "female": 60,
      "male": 40,
      "unit": "%",
      "source": "Table 1",
      "confidence": "High"
    }
  },
  "intervention_exposure": {
    "intervention_type": "Aerobic exercise",
    "details": "30 minutes of moderate-intensity exercise, 3 times per week",
    "duration": "12 weeks",
    "source": "Methods section, paragraph 3",
    "confidence": "High"
  },
  "comparison_control": {
    "control_type": "Usual care",
    "details": "Participants continued their normal daily activities",
    "source": "Methods section, paragraph 3",
    "confidence": "High"
  },
  "outcome_cognitive_function": {
    "baseline":{
      "intervention_group": {
        "mean": 27.5,
        "sd": 2.5,
        "unit": "MMSE score",
        "confidence_interval": {
           "95_percent": [25, 29],
           "type": "Wald",
           "unit": "points"
         },
        "source": "Table 1",
        "confidence": "High"
      },
      "control_group": {
        "mean": 25.0,
        "sd": 2.8,
        "unit": "MMSE score",
        "confidence_interval": {
           "95_percent": [22.0, 26.0],
           "type": "Wald",
           "unit": "points"
         },
        "source": "Table 1",
        "confidence": "High"
      }
    },
    "final_followup": {
      "time_point": "12 months",
      "intervention_group": {
        "mean": 28.5,
        "sd": 1.5,
        "unit": "MMSE score",
        "confidence_interval": {
           "95_percent": [27.5, 29.5],
           "type": "Wald",
           "unit": "points"
         },
        "source": "Table 2",
        "confidence": "High"
      },
      "control_group": {
        "mean": 27.0,
        "sd": 2.2,
        "unit": "MMSE score",
        "confidence_interval": {
           "95_percent": [26.0, 28.0],
           "type": "Wald",
           "unit": "points"
         },
        "source": "Table 2",
        "confidence": "High"
      }
    }
  },
  "outcome_depressive_symptoms": {
    "baseline": {
      "intervention_group": {
        "mean": 9.8,
        "sd": 2.1,
        "source": "Table 3",
        "confidence": "High"
      },
      "control_group": {
        "mean": 10.3,
        "sd": 1.1,
        "source": "Table 3",
        "confidence": "High"
      }
    },
  "final_followup": {
    "time_point": "12 months",
    "intervention_group": {
      "mean": 8.2,
      "sd": 2.1,
      "source": "Table 3",
      "confidence": "High"
    },
    "control_group": {
      "mean": 9.5,
      "sd": 1.3,
      "source": "Table 3",
      "confidence": "High"
    }
   }
  },
  "outcome_physical_function": {
    "baseline": {
    "intervention_group": {
      "mean": 52.7,
      "sd": 2.8,
      "source": "Table 1",
      "confidence": "High"
    },
    "control_group": {
      "mean": 46.3,
      "sd": 4.3,
      "source": "Table 1",
      "confidence": "High"
    }
    },
  "final_followup": {
    "time_point": "12 months",
    "intervention_group": {
      "mean": 55.7,
      "sd": 6.3,
      "source": "Table 5",
      "confidence": "Medium"
    },
    "control_group": {
      "mean": 50.3,
      "sd": 5.6,
      "source": "Table 5",
      "confidence": "Medium"
    }
   }
  },
  "eligibility_criteria": {
    "inclusion": ["Aged 65 or older", "No diagnosis of dementia"],
    "exclusion": ["Severe mobility limitations", "Uncontrolled cardiovascular disease"],
    "source": "Methods section, paragraph 1",
    "confidence": "High"
  },
  "adverse_events": {
    "total": 5,
    "serious": 1,
    "muscle_soreness": 3,
    "minor_falls": 2,
    "source": "Results section, paragraph 4",
    "confidence": "Medium"
  },
  "dropouts": {
    "total": 10,
    "intervention_group": 5,
    "control_group": 5,
    "source": "Results section, paragraph 4",
    "confidence": "Medium"
  },
  "pdf_status": "Processed",
  "notes": "Final follow-up was inferred from the study duration of 12 months. Extracted from 
  supplementary materials."
}
``` 
\end{verbatim}
\end{mdframed}

\section{Prompts}
\subsection{Self-Reflection Prompt}
\begin{mdframed}[backgroundcolor=gray!10]
\fontsize{8pt}{10pt}\selectfont
\begin{verbatim}
You are a world-leading expert in medical literature data extraction. Your task is to re-
evaluate the accuracy of the extracted data from the provided randomised controlled trial 
(RCT) PDF article, based on the JSON output you previously generated from the PDF to ensure 
it is suitable for meta-analysis, verifying correctness, completeness, and consistency and 
ensuring the data meets the highest standards of reliability and precision. Please ensure you 
are using the previously generated JSON output for further analysis. 

Task Requirements:
   - Your task is *binary*: Each extracted field is *either correct or incorrect*.  
   - Identify *ONLY the incorrect fields* in the original extraction. If the original value 
   is correct according to the PDF, DO NOT INCLUDE IT, even if you could add more detail or 
   context. The output JSON MUST ONLY contain entries for fields where the original value is 
   demonstrably wrong.
   - Provide a *corrected value* and a *justification* citing specific evidence from the PDF.

Self-Reflection and Reevaluation Steps: 
1. Iterate Through JSON: You MUST systematically iterate through every key-value pair in the 
initial JSON output, only record issues for fields with errors or inconsistencies in value, 
source, or confidence, except `"pdf_status"` and `"notes"`.
2. Validation Against Source Structure:  
   - Ensure that extracted values match the expected format based on study sections (e.g., 
   outcome measures should be from tables or results, not introduction).  
   - Identify any *structural inconsistencies* (e.g., missing key study characteristics, 
   incomplete sample size reporting).  
3. Internal Consistency Checks:
   - *Mathematical consistency*: Ensure numerical values are logically consistent (e.g., 
   total sample size = sum of intervention + control).  
   - *Unit consistency*: Verify all measurements use the correct units (e.g., blood pressure 
   should be in mmHg).  
   - *Data range validation*: Ensure extracted values fall within expected medical/clinical 
   ranges. 
4. Direct PDF Comparison: For *each key-value pair*: 
   - Review the extracted data alongside the original PDF content.
   - Identify discrepancies and data conflicts between different sections of the paper (e.g., 
   abstract vs. results section vs. tables).
5. Critical Assessment: For each key-value pair, consider these questions:
   - Accuracy: Does it exactly match the article’s reported data? 
   - Data Conflicts: Do different sections of the PDF report inconsistent values? 
   - Relevance: Is this data critical for meta-analysis?
   - Justification: Is the extracted value backed by direct evidence from the PDF?  
   - Source Appropriateness: Is the cited "source" the most appropriate location in the paper 
   for this data? If not, provide a more accurate source.
   - Confidence Justification: Is the assigned "confidence" level justified based on the 
   clarity and consistency of the data within the source and across the document? If not, 
   provide a revised confidence level ("High", "Medium", or "Low") and explain why.
6. Outcome Measures Verification:  
   - Confirm that all *time points* are correctly extracted and *final follow-up* is 
   prioritised.  
   - If the extracted value includes *median & IQR* or *range* and requires transformation to 
   calculate mean and standard deviation, ensure `"needs_transformation": true` is set. 
   Otherwise, ensure `"needs_transformation": true` is not set.
   - Ensure *p-values, confidence intervals, and effect sizes* are *exactly* as reported 
   (without rounding errors).
7. Adverse Events and Dropouts:  
   - Verify if *dropout data* or *adverse events* is extracted (if reported). 
8. Completeness Check: 
    - For fields marked as `"null"` or `"Not reported"` in the initial JSON, verify that this 
    status accurately reflects the information presented (or lack thereof) in the PDF. Do not 
    attempt to fill in missing values, maintain original status. 
9. Adjust or Skip:
   - If the initial extracted value is INCORRECT, include the following keys in the output: 
       - corrected and improved `"revised_value"`. Ensure that only the keys are changed that 
       there is a mismatch to the original extract. DO NOT include unchanged key-value pairs 
       from the original object.
       - justification citing the PDF that explain the `"revised_value"` changes.
       - If ONLY the source needs correction while the value remains unchanged, the output 
       MUST NOT include `"revised_value"` or `"initial_value"`, ONLY provide the corrected 
       `"revised_source"`. 
       - If ONLY the "confidence" were incorrect, ONLY provide the corrected 
       `"revised_confidence"`.
   - *Crucially*, if a dictionary or object (e.g., participant_characteristics, 
   outcome_cognitive_function) contains an error in only one of its sub-fields (e.g., age 
   within participant_characteristics), your `"revised_value"` and `"initial_value"` MUST 
   INCLUDE ONLY the *minimal* set of existing sub-fields necessary to provide context for 
   *identifying* the specific location of the error within the original JSON. For instance, 
   include parent keys and any identifying values (e.g., units, names, sources) that are 
   crucial for pinpointing the incorrect field.  Do NOT include any sub-fields that are 
   unrelated to identifying the error's location.
   - If the initial extraction is *CORRECT*, *DO NOT INCLUDE IT IN THE OUTPUT JSON*.
10. Include PDF Status:
   - Always include `"pdf_status"`: `"Processed"` or `"Unreadable"` in the output, even when 
   corrections are needed, to confirm the PDF was analysed.

Definition of "Error": An "error" means the originally extracted value is factually incorrect 
according to the PDF. It does not include cases where the original value is correct but could 
be more detailed or specific.

Minimal Change Principle: When correcting an error, only change the minimum number of keys 
necessary to fix the factual inaccuracy. Do not add extra information or details that were 
not part of the original extraction.

Output Format:
Provide your re-evaluation in a structured JSON format. The new output JSON MUST *ONLY* 
contain entries for the fields in the original JSON output that require correction. Correct 
fields MUST NOT appear in the output. For each field that needs correction, include the 
following:
`"field_name"`: The name of the field in the initial JSON that needs correction.
`"initial_value"`: The originally extracted value from the initial JSON output that is 
incorrect.
`"revised_value"`: The corrected value, containing only the keys that differ from the initial 
extraction. Do not include unchanged key-value pairs from the original object.
`"justification"`: A concise explanation of why the original value was incorrect and why the 
revised value is correct, citing specific evidence from the PDF.
`"revised_source"`: (Optional) If the original "source" was incorrect AND the 
`"initial_value"` isn't changed, ONLY provide the corrected source (the output MUST NOT 
include `"revised_value"` or `"initial_value"`); if the extracted data requires changes 
please do include with `"revised_value"`.
`"revised_confidence"`: (Optional) If ONLY the confidence level has a mismatch of reliability 
from the initial extraction, provide one of the following confidence level ("High", "Medium", 
or "Low") and a justification of the mismatch, if the extracted data requires changes please 
do include with revised value, include with `"revised_value"`.
Otherwise, if NO corrections are needed, respond with: 
```json
{"status": "No corrections needed", "pdf_status": "Processed" }.
```

Example of Input JSON: 
```json
{
  "title": "The Impact of Regular Exercise on Cognitive Function in Older Adults",
  "first_author": "Jane Doe",
  "publication_year": 2023,
  "country": "USA",
  "study_design": {
      "randomisation": "Randomised assignment to intervention or control group",
      "blinding": "Double-blinded",
      "allocation_concealment": "Adequate",
      "study_duration": "12 months",
      "source": "Methods section, paragraph 2",
      "confidence": "High"
  },
  "sample_size": {
    "total": 100,
    "intervention_group": 50,
    "control_group": 50,
    "source": "Table 1",
    "confidence": "High"
  },
  "study_characteristics": {
    "study_setting": "Community center",
    "funding_source": "National Institutes of Health",
    "ethical_approval": "Institutional Review Board approved",
    "source": "Methods section, paragraph 1",
    "confidence": "High"
  },
  "participant_characteristics": {
    "age": {
      "range": [22,57],
      "unit": "years",
      "source": "Table 1",
      "confidence": "High"
    },
    "gender_distribution": {
      "female": 60,
      "male": 40,
      "unit": "%",
      "source": "Table 1",
      "confidence": "High"
    }
  },
  "intervention_exposure": {
    "intervention_type": "Aerobic exercise",
    "details": "30 minutes of moderate-intensity exercise, 3 times per week",
    "duration": "12 weeks",
    "source": "Methods section, paragraph 3",
    "confidence": "High"
  },
  "comparison_control": {
    "control_type": "Usual care",
    "details": "Participants continued their normal daily activities",
    "source": "Methods section, paragraph 3",
    "confidence": "High"
  },
  "outcome_cognitive_function": {
    "primary_outcome": true,
    "final_followup": {
      "time_point": "12 months",
      "intervention_group": {
        "mean": 28.5,
        "sd": 1.5,
        "unit": "MMSE score",
        "confidence_interval": {
           "95_percent": [27.5, 29.5],
           "type": "Wald",
           "unit": "points"
         },
        "p_value": 0.04,
        "source": "Table 2",
        "confidence": "High"
      },
      "control_group": {
        "mean": 27.0,
        "sd": 2.2,
        "unit": "MMSE score",
        "confidence_interval": {
           "95_percent": [26.0, 28.0],
           "type": "Wald",
           "unit": "points"
         },
        "p_value": 0.04,
        "source": "Table 2",
        "confidence": "High"
      }
    },
    "source": "Table 2",
    "confidence": "High"
  },
  "eligibility_criteria": {
    "inclusion": ["Aged 65 or older", "No diagnosis of dementia"],
    "exclusion": ["Severe mobility limitations", "Uncontrolled cardiovascular disease"],
    "source": "Methods section, paragraph 1",
    "confidence": "High"
  },
  "adverse_events": {
    "total": 5,
    "serious": 1,
    "muscle_soreness": 3,
    "minor_falls": 2,
    "source": "Results section, paragraph 4",
    "confidence": "Medium"
  },
  "dropouts": {
    "total": 10,
    "intervention_group": 5,
    "control_group": 5,
    "source": "Results section, paragraph 4",
    "confidence": "Medium"
  },
  "pdf_status": "Processed",
  "notes": "Final follow-up was inferred from the study duration of 12 months. Extracted from 
  supplementary materials."
}
```

Your output *MUST* strictly follow this JSON format:
Example1: Reflection Output (Corrections Needed):
```json
{
  "pdf_status": "Processed",
  "data_corrections": [
   {
      "field_name": "participant_characteristics",
      "initial_value": {
        "age": {
          "range": [22,57]
        }
      },
      "revised_value": {
        "age": {
          "range": [21,57]
        }
      },
      "justification": "The range for age was incorrect. The correct range from Table 1 is 
      (21-57).",
      "revised_source": "Table 1"
    },
   {
      "field_name": "outcome_cognitive_function",
      "initial_value": {
         "control_group": {
            "mean": 27.0,
          }
        },
      "revised_value": {
          "control_group": {
            "mean": 26.0,               
          }
        },
      "justification": "Re-evaluated outcome measures with Table 4, the mean for the control 
      group was 27, fixed to 26",
      "revised_source": "Table 4",
      "revised_confidence": "High"
  }
  ]
}
```
Example2: Reflection Output (No Corrections Needed):
```json
{"status": "No corrections needed", "pdf_status": "Processed" }.
```

\end{verbatim}
\end{mdframed}

\subsection{Combined EXT Prompt}
\begin{mdframed}[backgroundcolor=gray!10]
\fontsize{8pt}{10pt}\selectfont
\begin{verbatim}
You are a word-leading data analysis expert for meta-analysis. Below are three JSON outputs 
generated by different large language models, all extracting data from the same RCTs article. 
The JSON structure is consistent, but field values may differ slightly between models.

Your task is to merge these three JSONs into a single, unified, and accurate version by 
following the rules below.

Merging Rules:
1. For each field:
   - If two models agree on a value and the third differs, use the majority (2-vs-1 voting).
   - If all three values are the same, keep it as is.
   - If all three values are different:
     - If a "confidence" field is present, choose the one with the highest confidence 
     (prioritizing "High" > "Medium" > "Low").
     - Otherwise, choose the version with the most complete, consistent, and logically sound 
     information.
       - Completeness: The version with the most populated fields within the relevant nested 
       field.
       - Consistency: The version where data types align with expected types (e.g., numerical 
       values are numbers). Resolve obvious inconsistencies where possible.
2. For nested fields (such as `outcome_bmd`, `participant_characteristics`, etc.), apply the 
same rules recursively.
3. Maintain the original JSON structure in the final result.
4. Do NOT include any explanations or commentary. Just return the final merged JSON object.

Please return the final merged JSON.

---

### Model A Output:
<PASTE FULL JSON OF MODEL A HERE>


### Model B Output:
<PASTE FULL JSON OF MODEL B HERE>


### Model C Output:
<PASTE FULL JSON OF MODEL C HERE>


\end{verbatim}
\end{mdframed}

\subsection{Customised EXT Prompt}

\begin{mdframed}[backgroundcolor=gray!10]
\begin{HighlightVerbatim}
You are a world-leading expert in \textcolor{red}{orthopedic and metabolic bone disease} literature data 
extraction. Your task is to extract structured research data from RCT PDFs to enable
meta-analysis. The extracted data must be formatted precisely, ensuring alignment with 
meta-analysis requirements.

Input: A full-text research PDF document.

Task: Extract study-related information using a structured approach
   - Study Characteristics (SC): General study details (e.g., study setting, funding source, 
   ethical approval).
   - Participant Characteristics (PC): Extract demographic and clinical baseline information 
   separately for intervention and control groups, if available 
   \textcolor{red}{(e.g., age, gender distribution, BMI)}. Maintain the original structure of the source and do 
   not merge or average across groups.
   - Intervention/Exposure (IE): Treatment, exposure, or intervention details (e.g., type, 
   dosage, frequency, duration of intervention).
   - Comparison/Control (CC): Description of the control/comparison group (e.g., usual care, 
   placebo).
   - Outcome Measures (OM): Extract all reported outcomes with their values, including time 
   points and statistics. \textcolor{red}{Please Focus on these outcomes: Bone Mineral Density (femoral neck, }
   \textcolor{red}{total hip, lumbar spine) and Bone Turnover Markers (CTX, P1NP, BONE ALP, Osteocalcin).}
   - Study Design (SD): Methodology details (e.g., randomisation, blinding [single, double, 
   triple?], allocation concealment, study duration). Pay special attention to aspects of 
   study design relevant to bias assessment.

......
......
(\textcolor{blue}{The rest of the prompt are the same as the baseline extraction prompt})

\end{HighlightVerbatim}
\end{mdframed}

\subsection{Evaluation Prompts}

\subsubsection{Statistical Evaluation Prompt}

\begin{mdframed}[backgroundcolor=gray!10]
\fontsize{8pt}{10pt}\selectfont
\begin{verbatim}
You are a highly skilled data evaluator for a meta-analysis project, specializing in the 
assessment of complex statistical data extracted from research papers. Your primary task is 
to compare statistical information extracted by a large language model (EXT) with a gold-
standard, human-annotated ground truth (GT), both provided in structured JSON format.

The data may include nested fields and meta-information such as confidence, source, and 
notes, which are provided **only for context** and **MUST** not be treated as primary values 
unless otherwise stated.

STEP-BY-STEP INSTRCTIONS:
Step 1: Field Matching
For each relevant statistical field present in the **GT JSON**, diligently attempt to 
identify the corresponding field in the **EXT JSON**. Prioritize the following matching 
strategies, in order:
1.  *Exact Name Match:* If a field with precisely the same name exists in both GT and EXT, 
consider it a direct match. This is the preferred method.
2.  *Semantic Similarity Match:* If an exact match is *not* found, use your expert knowledge 
of statistical terminology to identify fields with similar *meaning*.  Consider variations 
in naming conventions.  For example:
    - `"LGL_group"` is likely semantically equivalent to `"intervention_group"` or 
    `"treatment_group"`
    - `"Mean_Difference"` is likely semantically equivalent to `"Difference_in_Means"`
    - `"Standard Deviation"` is likely semantically equivalent to `"SD"` or `"sd"`
    *If semantic similarity is HIGH but you are uncertain, carefully examine any notes or 
    context surrounding the extracted value (see Step 2).*
3.  *Missing Field:* If a field exists in GT but is not found in EXT (after exact and 
semantic matching), you must mark it as `"Missing"` and count it as a False Negative.
**Note:** EXT fields may be nested. You must traverse the full structure to find possible 
semantic matches.

**Step 2: Value Comparison and Meta-Information Assessment**
For each GT field for which a matching EXT field has been identified, perform a detailed 
value comparison and assessment of any available meta-information.  Use the following 
guidelines:
1.  **Numerical Values:**
    -   A numerical value in EXT is considered "Correct" if it falls within ±1% of the 
    corresponding numerical value in GT.  Calculate percentage difference as: `abs(EXT_value 
    - GT_value) / GT_value`.
    -   **Units:**  The extracted value must be expressed in *equivalent* units or properly 
    convertible. For example:
        -  `"kg/m²"` and `"kg/m^2"` are considered equivalent; `"count"` and `"n"` are 
        considered equivalent.
        -  `"grams"` and `"kilograms"` are *not* considered equivalent without appropriate 
        conversion. Attempt conversion if possible and well-defined; otherwise mark as 
        "Hallucinated" with error_type: `"Incorrect unit"`.
    If EXT gives partial information (e.g., mean but no SD), you may still mark it "Correct" 
    if GT doesn’t expect the missing part. Otherwise, explain.
    -   **Meta-Information**: Prioritize values with higher confidence and more reliable 
    sources. If a value is flagged as having low confidence or originating from a less 
    reliable source (e.g., a figure legend instead of the main text), carefully scrutinize 
    its accuracy.  Use notes to describe the source.
2.  **String Values:**
    -  A string in EXT is considered **"Correct"** if its **meaning** is semantically 
    equivalent to the GT value. Do **not** require exact character matches.
    You should:
        - Ignore case, formatting, hyphens, extra whitespace
        - Accept rewordings if meaning is unchanged
        - Evaluate synonym phrases as equivalent
    Examples of equivalent strings:
        - `"low glycemic load diet"` equivalent to `"Low-GL dietary"`
        - `"6-month follow-up"` equivalent to `"follow-up(6 months)"`
Always provide a brief justification if a string is semantically matched or deemed 
different. If meaning is different, mark as `"Hallucinated"` or `"Overgeneralized"` depending 
on content loss
3. **Special Case: "null" or "Not reported" Values in EXT**
EXT fields may contain: `"null"` for missing numerical values or `"Not reported"` or 
equivalent words for missing non-numerical values. This reflects that the extraction system 
could not find these values in the original PDF (as per extraction prompt instructions).
    - If GT expects a value but EXT gives null / "Not reported", which is not "Incorrect", 
    *BUT* it must be marked as "Missing" (i.e., a False Negative). 
    - If GT does not include the field: *DO NOT* evaluate or penalize this EXT field.

**Step 3: Field Evaluations**
For each GT field, output:
```json
{
  "field_name": "GT field name (or full path)",
  "status": "Correct | Hallucinate | Missing",
  "error_type": "Only required when status is Hallucinate or Missing",
  "explanation": "Only required when status is Hallucinate or Missing"
}
```
If "status" is "Correct", do not include "error_type" or "explanation" in the output.
Only include "error_type" and "explanation" when "status" is "Hallucinate" or "Missing".

Use one of the following for "error_type" only when status is "Hallucinate" or "Missing":
  - "Missing field": A required data item is completely absent from the extracted output but 
  present in the ground truth.
  - "Incorrect value": The extracted field is present but its value does not match the 
  ground truth (e.g., numerical or textual mismatch).
  - "Incorrect unit": The extracted value is correct in magnitude but the unit is wrong or 
  inconsistent with the ground truth (e.g., "minutes" instead of "hours"). 
  - "Overgeneralized": The extracted information is broader or less specific than the ground 
  truth, losing important qualifying details (e.g., applying a subgroup result to the entire 
  population).



---
## GT FIELDS (Ground Truth)

<!-- GT_INSERT -->


## EXT FIELDS (Extracted)

<!-- EXT_INSERT -->

\end{verbatim}
\end{mdframed}

\subsubsection{Quality Assessment Evaluation Prompt}

\begin{mdframed}[backgroundcolor=gray!10]
\fontsize{8pt}{10pt}\selectfont
\begin{verbatim}
You are a highly skilled data evaluator for a meta-analysis project, specializing in the 
assessment of study quality extracted from randomised controlled trials (RCTs). Your task is 
to rigorously compare quality-related fields extracted by a large language model (EXT) with 
a gold-standard, human-annotated ground truth (GT). Both are provided to you in structured 
JSON format. Your evaluation should prioritize semantic meaning over exact text match, as 
quality items are often expressed with variable language. Your evaluation should prioritize 
**semantic meaning** over exact word match, as quality fields are often expressed in varied 
language.

Please follow the steps below:

**Input:** 
- A GT JSON containing expected study quality fields  
- An EXT JSON containing extracted values  
- EXT may include meta-information such as: `confidence`, `source`, `notes`


**Step 1: Field Matching**
For each field in the **GT JSON**, identify the matching field in the **EXT JSON**, using 
the following strategy:
1. *Exact Match:* Use this when both field names are identical.
2. *Semantic Match:* Match fields with equivalent *meaning*, even if the names differ. For 
example:
   - `"randomised controlled trial"` equivalent to `"randomised"`
   - `"blinding of outcome assessors"` equivalent to `"outcome assessor blinded"`
   - `"ethics approved"` equivalent to `"approved by an institutional review board"`
3. *Missing Field:* If no semantically equivalent field exists in EXT, mark the GT field as 
"Missing".

You may refer to EXT field meta-information (e.g., `source`, `notes`, `confidence`) to aid 
in field matching, especially when EXT uses vague or ambiguous labels.


**Step 2: Value Comparison and Meta Evaluation**
For each matched field, assess the *semantic correctness* of the extracted value. Use the 
following guidelines:
1. **Correct:** if
   - The EXT value expresses the *same meaning* as GT (even if phrasing differs)
   - Examples:
       - `"not reported"` equivalent to `"not mentioned"`
       - `"randomly assigned"` equivalent to `"randomised"`
       - `"IRB approved"` equivalent to `"ethics approval obtained"`

2. **Incorrect Values:**
   - The EXT value has a *different meaning* than the GT
   - The value misrepresents study design or mislabels methods

3. **Special Case: "null" or "Not reported" Values in EXT**
EXT fields may contain: `"Not reported"` or equivalent words for missing values. This 
reflects that the extraction system could not find these values in the original PDF (as per 
extraction prompt instructions).
    - If GT expects a value but EXT gives "Not reported", which is **not "Incorrect"**, 
    *BUT* it must be marked as "Missing" (i.e., a False Negative). Explanation should note: 
    `"EXT marked as Not reported"`
    - If GT does not include the field: *DO NOT* evaluate or penalize this EXT field.


## Step 3: Field Evaluation Output 
For each GT field, output a valid, completed JSON object with:

```json
{
  "field_name": "GT field name (or full path)",
  "status": "Correct | Hallucinated | Missing",
  "error_type": "NOT REQUIRED for Correct, REQUIRED for others",
  "explanation": "Short justification (mandatory for Hallucinated, not required for 
  Correct)"
}
```
Use one of the following for "error_type":
  - "Missing field": A required data item is completely absent from the extracted output but 
  present in the ground truth.
  - "Incorrect value": The extracted field is present but its value does not match the 
  ground truth (e.g., numerical or textual mismatch).
  - "Incorrect unit": The extracted value is correct in magnitude but the unit is wrong or 
  inconsistent with the ground truth (e.g., "minutes" instead of "hours"). 
  - "Overgeneralized": The extracted information is broader or less specific than the ground 
  truth, losing important qualifying details (e.g., applying a subgroup result to the entire 
  population).


---
## GT FIELDS (Ground Truth)

<!-- GT_INSERT -->


## EXT FIELDS (Extracted)

<!-- EXT_INSERT -->

\end{verbatim}
\end{mdframed}

\subsubsection{Study Information Evaluation Prompt}

\begin{mdframed}[backgroundcolor=gray!10]
\fontsize{8pt}{10pt}\selectfont
\begin{verbatim}
You are a highly skilled data evaluator for a meta-analysis project. Your task is to 
evaluate the correctness of **study-level metadata** extracted from research papers. 
Specifically, you will compare metadata extracted by a large language model (EXT) with a 
human-annotated ground truth (GT), both provided in JSON format.
This evaluation focuses on **string-level accuracy**, allowing for **minor formatting and 
typographical variations**. Your goal is to assess whether the EXT output meaningfully 
matches the GT values.

The data may include nested fields and meta-information such as confidence, source, and 
notes, which are provided **only for context** and **MUST** not be treated as primary values 
unless otherwise stated.

STEP-BY-STEP INSTRCTIONS:
**Step 1: Field Matching**
For each relevant study information field present in the **GT JSON**, diligently attempt to 
identify the corresponding field in the **EXT JSON**. Prioritize the following matching 
strategies, in order::
1.  *Exact Name Match:* If a field with precisely the same name exists in both GT and EXT, 
consider it a direct match. This is the preferred method.
2.  *Semantic Similarity Match:* If an exact match is *not* found, use your expert knowledge 
of study information terminology to identify fields with similar *meaning*.  Consider 
variations in naming conventions.  For example:
    - `"study_characteristics.PC"` is likely semantically equivalent to 
    `"participant_characteristics"`
    *If semantic similarity is HIGH but you are uncertain, carefully examine any notes or 
    context surrounding the extracted value (see Step 2).*
3.  *Missing Field:* If a field exists in GT but is not found in EXT (after exact and 
semantic matching), you must mark it as `"Missing"` and count it as a False Negative.
**Note:** GT and EXT fields may be nested. You must traverse the full structure to find 
possible semantic matches. You may refer to EXT field meta-information (e.g., `source`, 
`notes`, `confidence`) to aid in field matching, especially when EXT uses vague or ambiguous 
labels.

**Step 2: Value Comparison**
1. A field is *Correct* if:
    - The EXT value expresses the *same meaning* as GT (even if phrasing differs)
    - Ignore case, formatting, hyphens, extra whitespace
    - Accept rewordings if meaning is unchanged
    - Evaluate synonym phrases as equivalent
2. A field is *Hallucinated* if:
    - There is a meaningful content mismatch. If meaning is different, mark as `"Hallucinated"` 
    or `"Overgeneralized"` depending on content loss
3. **Special Case: "Not reported" Values in EXT**
EXT fields may contain: `"Not reported"` or equivalent words for missing values. This 
reflects that the extraction system could not find these values in the original PDF (as per 
extraction prompt instructions).
    - If GT expects a value but EXT gives "Not reported", which is **not "Incorrect"**, 
    *BUT* it must be marked as "Missing" (i.e., a False Negative). Explanation should note: 
    `"EXT marked as Not reported"`
    - If GT does not include the field: *DO NOT* evaluate or penalize this EXT field.


**Step 3: Field Evaluations**
For each GT field, output:
```json
{
  "field_name": "GT field name (or full path)",
  "status": "Correct | Hallucinated | Missing",
  "error_type": "NOT REQUIRED for Correct, REQUIRED for others",
  "explanation": "Short justification (mandatory for Hallucinated and Missing, not required for 
  Correct)"
}
```
Use one of the following for "error_type":
  - "Missing field": A required data item is completely absent from the extracted output but 
  present in the ground truth.
  - "Incorrect value": The extracted field is present but its value does not match the 
  ground truth (e.g., numerical or textual mismatch).
  - "Incorrect unit": The extracted value is correct in magnitude but the unit is wrong or 
  inconsistent with the ground truth (e.g., "minutes" instead of "hours"). 
  - "Overgeneralized": The extracted information is broader or less specific than the ground 
  truth, losing important qualifying details (e.g., applying a subgroup result to the entire 
  population).



Return only a **valid JSON object**. Do not include markdown formatting, explanation, or 
code blocks.

---
## GT FIELDS (Ground Truth)

<!-- GT_INSERT -->



## EXT FIELDS (Extracted)

<!-- EXT_INSERT -->

\end{verbatim}
\end{mdframed}

\subsection{Error Distribution Across Fields}
To better understand model failures, we examined how error types were distributed across different fields. Table \ref{tab:field-errors} shows the fields with the highest error rates. The field groups in this study are:
\begin{itemize}
    \item Study Characteristics (SC): general study metadata.
    \item Participant Characteristics (PC): demographic and baseline participant details.
    \item Intervention/Exposure (IE): information about treatments or exposures studied.
    \item Outcome Measures (OM): primary and secondary outcome variables.
    \item Study Design (SD): trial design, randomisation, blinding, and related details.
    \item Adverse Events (AE): safety outcomes, including total, serious events or dropouts.
\end{itemize}

As shown in Table \ref{tab:field-errors}, error types vary across fields. In OM, most errors were missing fields (95.3\%), indicating that models often fail to identify outcome variables, likely due to their varied expression or placement (e.g., within tables or result summaries). In contrast, PC had a higher incorrect value errors (24.6\%), pointing to frequent mistakes in misinterpreting group-specific details (e.g., age, BMI, sex distribution). The IE field showed the highest rate of overgeneralization errors (28.6\%), where models often missed key subgroup-specific treatment details, leading to overly general or incomplete summaries compared to the ground truth. SD also had a noticeable overgeneralization rate (4.1\%) alongside incorrect value errors (23.6\%), reflecting challenges in interpreting ambiguous reports of design elements like blinding or randomisation. Meanwhile, SC and AE were primarily affected by missing field errors (86.5\% and 95.5\%, respectively), with some factual errors. These results indicate that fields with structured numerical data, such as PC and OM, often have missing or incorrect values, while narrative fields, like IE and SD, are more likely to show overgeneralization errors. 

{
\begin{table}[h]
\centering
\small
\begin{tabular}{lrrrr}
\toprule
\textbf{Field} & \textbf{Missing} & \textbf{Incorrect Value} & \textbf{Incorrect Unit} & \textbf{Overgeneralized} \\
\midrule
Outcome Measures (OM) & 95.2\% (14{,}497) & 4.3\% (660) & 0.5\% (78) & 0.0\% (0) \\
Participant Characteristics (PC) & 72.8\% (2{,}522) & 24.6\% (853) & 2.5\% (85) & 0.2\% (6) \\
Study Design (SD) & 71.1\% (367) & 23.6\% (122) & 0.0\% (0) & 5.2\% (27) \\
Study Characteristics (SC) & 86.5\% (372) & 13.5\% (58) & 0.0\% (0) & 0.0\% (0) \\
Intervention/Exposure (IE) & 64.5\% (142) & 6.8\% (15) & 0.0\% (0) & 28.6\% (63) \\
Adverse Events (AE) & 95.5\% (461) & 3.7\% (30) & 0.0\% (0) & 0.2\% (1) \\
\bottomrule
\end{tabular}
\caption{Top Error type distribution across field groups}
\label{tab:field-errors}
\end{table}
}

\subsection{Error Distribution Across Models with Studies}
In order to examine model-specific weakness, we analysed how extraction errors were distributed across combinations of model and studies datasets. Table \ref{tab:model_errors} highlight the five combinations with the highest error counts for each model, showing both the total errors and their breakdown by type. GPT had the highest error counts in most combinations, particularly for OM in studies like MA6 (2,377 errors, 97.5\% missing) and MA2 (1,323 errors, 98.5\% missing). This shows GPT often misses outcome-related variables. In contrast, Gemini showed a more balanced error distribution, with fewer total errors but a wider range of error types. For instance, in OM for MA6, missing fields accounted for 89.4\% of errors, while incorrect values made up 6.1\% (79 errors) and incorrect units 2.5\% (32 errors), which means Gemini attempts more extraction but struggles with numerical accuracy. Grok also had high missing field rates, such as 94.4\% (1,976 errors) in OM for MA6. In OM for MA2, Grok had a higher incorrect value errors (17.7\%, 104 errors). Notably, it also produced the relatively fewest errors in the PC field, suggesting stronger performance in handling demographic and baseline variables. Across all models, the OM field presented the greatest challenges, mainly due to high rates of missing information. Analysis of error distribution highlights different models behaviours: GPT takes a cautious approach, often skipping content; Gemini attempts more comprehensive extractions but struggles with numerical precision; and Grok had the fewest errors in PC among the three models.

{
\begin{table}[ht]
\centering
%\small
\fontsize{8pt}{10pt}\selectfont
\begin{tabular}{lccccccc}
\toprule
\textbf{Model} & \textbf{Field} & \textbf{Studies} & \textbf{Missing} & \textbf{Incorrect Value} & \textbf{Incorrect Unit} & \textbf{Overgeneralized}  \\
\midrule
\multirow{5}{*}{GPT}
& OM & MA6 & 97.5\% (2377) & 2.3\% (56) & 0.2\% (4) & 0.0\% (0) \\
& OM & MA2 & 98.5\% (1323) & 1.5\% (20) & 0.0\% (0) & 0.0\% (0) \\
& OM & MA5 & 98.0\% (1050) & 2.0\% (21) & 0.0\% (0) & 0.0\% (0) \\
& OM & MA3 & 95.9\% (674) & 4.1\% (29) & 0.0\% (0) & 0.0\% (0) \\
& PC & MA6 & 55.8\% (154) & 37.0\% (102) & 7.2\% (20) & 0.0\% (0) \\
\midrule
\multirow{5}{*}{Gemini}
& OM & MA6 & 89.4\% (1150) & 6.1\% (79) & 2.5\% (32) & 0.0\% (0) \\
& OM & MA2 & 96.7\% (741) & 3.3\% (25) & 0.0\% (0) & 0.0\% (0) \\
& OM & MA5 & 95.7\% (633) & 2.3\% (15) & 0.6\% (4) & 0.0\% (0) \\
& OM & MA3 & 92.7\% (408) & 6.8\% (30) & 0.5\% (2) & 0.0\% (0) \\
& PC & MA1 & 72.5\% (232) & 23.4\% (75) & 1.3\% (4) & 0.3\% (1) \\
\midrule
\multirow{5}{*}{Grok}
& OM & MA6 & 94.4\% (1976) & 4.8\% (101) & 0.8\% (16) & 0.0\% (0) \\
& OM & MA5 & 91.3\% (717) & 8.7\% (68) & 0.0\% (0) & 0.0\% (0) \\
& OM & MA3 & 97.4\% (589) & 2.6\% (16) & 0.0\% (0) & 0.0\% (0) \\
& OM & MA2 & 82.3\% (482) & 17.7\% (104) & 0.0\% (0) & 0.0\% (0) \\
& PC & MA1 & 55.2\% (90) & 41.1\% (67) & 6.1\% (10) & 1.2\% (2) \\
\bottomrule
\end{tabular}
\caption{Top 5 error combinations per model across field and task}
\label{tab:model_errors}
\end{table}
}

%\end{comment}

\end{document}